\documentclass[lettersize,journal]{IEEEtran}
\usepackage{amsmath,amsfonts}
\usepackage{textcomp}
\usepackage{stfloats}
\usepackage{url}
\usepackage{verbatim}

\usepackage{xcolor,soul,framed} %,caption
\usepackage{booktabs}
\usepackage{subfigure}
\colorlet{shadecolor}{yellow}
\usepackage{cite}
\usepackage{array}
\usepackage{mdwmath}
\usepackage{mdwtab}
\usepackage{eqparbox}
\usepackage{url}

\usepackage{graphicx}
\usepackage[latin1]{inputenc}
\usepackage{multirow}
\usepackage{amssymb,amsmath}
\usepackage{algorithm}
\usepackage{algorithmic}

\newcommand{\bc}{\mathbf{c}}

\newcommand{\bp}{\mathbf{p}}

\newcommand{\bu}{\mathbf{u}}
\newcommand{\bw}{\mathbf{w}}

\newcommand{\balpha}{\boldsymbol{\alpha}}
\newcommand{\bbeta}{\boldsymbol{\beta}}
\newcommand{\bgamma}{\boldsymbol{\gamma}}

\newcommand{\Ltrain}{\mathcal{L}_{train}}
\newcommand{\Lval}{\mathcal{L}_{val}}
\newcommand{\Lvalp}{\mathcal{L}_{val'}}

\begin{document}
\bstctlcite{IEEEexample:BSTcontrol}
    \title{ZO-DARTS++: An Efficient and Size-Variable Zeroth-Order Neural Architecture Search Algorithm}
  \author{Lunchen~Xie,~\IEEEmembership{Student Member,~IEEE,}
      Eugenio~Lomurno,~\IEEEmembership{Student Member,~IEEE,}\\
      Matteo~Gambella,~\IEEEmembership{Student Member,~IEEE,}
      Danilo~Ardagna,~\IEEEmembership{Senior Member,~IEEE,}
      Manual~Roveri,~\IEEEmembership{Senior Member,~IEEE,}
      Matteo~Matteucci,~\IEEEmembership{Member,~IEEE,}
      ~and~Qingjiang~Shi,~\IEEEmembership{Member,~IEEE}% <-this % stops a space
      
  \thanks{This work has been submitted to the IEEE for possible publication. Copyright may be transferred without notice, after which this version may no longer be accessible.}
  \thanks{L. Xie and Q. Shi are with the School of Computer Science and Technology, Tongji University, Shanghai, China (e-mail: lcxie@tongji.edu.cn; shiqj@tongji.edu.cn). Q. Shi is also with Shenzhen Research Institute of Big Data, Shenzhen, China.}%
  \thanks{E. Lomurno, M. Gambella, D. Ardagna, M. Roveri, and M. Matteucci are with the Dipartimento di Elettronica, Informazione e Bioingegneria (DEIB), Politecnico di Milano, Milan, Italy (e-mail: eugenio.lomurno@polimi.it; matteo.gambella@polimi.it; danilo.ardagna@polimi.it; manuel.roveri@polimi.it; matteo.matteucci@polimi.it).}% <-this % stops a space
  \thanks{L. Xie is also with the DEIB, Politecnico di Milano, Milan, Italy. The main part of this work was completed by him as a visiting Ph.D. student, supported by the China Scholarship Council. (Corresponding authors: D. Ardagna and Q. Shi.)}
  }

\markboth{Journal of \LaTeX\ Class Files,~Vol.~XX, No.~XX, February~2025}%
{Xie \MakeLowercase{\textit{et al.}}: ZO-DARTS++: An Efficient and Size-Variable Zeroth-Order Neural Architecture Search Algorithm}

% ====================================================================
\maketitle

% === ABSTRACT ====================================================================
% =================================================================================
\begin{abstract}
Differentiable Neural Architecture Search (NAS) provides a promising avenue for automating the complex design of deep learning (DL) models. However, current differentiable NAS methods often face constraints in efficiency, operation selection, and adaptability under varying resource limitations. We introduce ZO-DARTS++, a novel NAS method that effectively balances performance and resource constraints. By integrating a zeroth-order approximation for efficient gradient handling, employing a sparsemax function with temperature annealing for clearer and more interpretable architecture distributions, and adopting a size-variable search scheme for generating compact yet accurate architectures, ZO-DARTS++ establishes a new balance between model complexity and performance. In extensive tests on medical imaging datasets, ZO-DARTS++ improves the average accuracy by up to 1.8\% over standard DARTS-based methods and shortens search time by approximately 38.6\%. Additionally, its resource-constrained variants can reduce the number of parameters by more than 35\% while maintaining competitive accuracy levels. Thus, ZO-DARTS++ offers a versatile and efficient framework for generating high-quality, resource-aware DL models suitable for real-world medical applications.
\end{abstract}

% === KEYWORDS ====================================================================
% =================================================================================

% \fi
\IEEEpeerreviewmaketitle

% ====================================================================
% ====================================================================
% ====================================================================

% === I. INTRODUCTION =============================================================
% =================================================================================
\section{Introduction}
\IEEEPARstart{E}{ngaged} in the era of artificial intelligence, deep learning (DL) methods have penetrated daily life and demonstrated powerful capabilities in all aspects and relevant tasks, such as self-driving\cite{ramos2017detecting}, medical diagnosis\cite{zhang2021cross}, and automatic chatbots\cite{yan2018chitty}. Technically speaking, AI applications are firmly based on various DL models, which include mainly two parts: parameters and structures. There has been a long period of effective application of optimization algorithms for training complicated DL models, which allows fast and accurate generation of optimal task-specific model parameters for different fields. In contrast, the design of model structures is a complex task. Thanks to concentrated efforts over a long period, numerous outstanding DL models emerged endlessly, such as VGG\cite{2014Very}, ResNet\cite{he2016deep}, DenseNet\cite{huang2017densely}, etc. These powerful models have become important components and references for DL model design nowadays. Regrettably, the design process is challenging to formulate by functions, unlike model parameters. Traditionally, it has relied heavily on a trial-and-error approach, demanding substantial domain expertise and considerable human effort. Meanwhile, the significant heterogeneity of data necessitates domain-specific adaptation of models. Hence, the need to automate and simplify the design process and to reduce human involvement arises.

To alleviate the burden of repetitive and demanding tasks for humans, the Neural Architecture Search (NAS) paradigm has been proposed. Today, it is considered a promising approach for automating the model design process~\cite{nasframework}. 

NAS aims to identify the optimal DL model for a given dataset without requiring human labor or expertise. Broadly, NAS methods can be categorized based on their search strategies: evolutionary algorithms (EA), reinforcement learning (RL), and differentiable NAS. The first two approaches search through a large number of candidate model architectures, with model parameters often fine-tuned during the search process to evaluate performance. The time and memory consumption during this process cannot be ignored even if the training is limited to small datasets. To address these challenges, differentiable NAS methods have been proposed, particularly those that train a supernet capable of encompassing all possible sub-structures. Among these, the most prominent method refers to DARTS~\cite{liu2018darts}, which is a pillar of weight-sharing approaches. By introducing a continuous structure weights, DARTS transforms the discrete search process into a continuous one, which is solvable with gradient-based optimization algorithms and greatly reduces search costs compared to EA- and RL-based methods. DARTS has received widespread attention and has spawned a large number of related algorithms to gain success in various tasks~\cite{xiesnas, dong2019searching, dong2019one, chu2020fair, zhang2021idarts}.

Despite the great achievements DARTS has made, there are still some troublesome issues. First, the search process in DARTS is cast as a bi-level optimization problem. This problem is intractable due to its special  structure and high dimensionality of the gradients~\cite{tsaknakis2022implicit}. DARTS solves this problem by approximating the implicit gradient trivially and sacrificing precision,  leaving room for further improvement. Second, the architecture generation process is not plausible~\cite{chu2020fair}. For a specific model layer, DARTS-style methods handle various operation candidates parallelly by assigning probability weights to them and evaluating the weighted sum of operations. When the search is complete, the operation corresponding to the highest probability is selected, equal to setting the final weight to 1 for that operation (and 0 for the rest). However, these methods often produce probabilities with low divergence, making the operation selection process debatable. In practice, DARTS often tends to assign larger probabilities for skip connections. Thus, the final model has only non-learnable operations, which may perform badly~\cite{liang2019darts+, xie2023zo}. The last issue is the improper handling of resource consumption. Although differentiable NAS methods can easily include resource constraints~\cite{jin2019rc}, they focus only on penalizing the optimization process rather than designing novel structures and are therefore not flexible enough. In conclusion, DARTS-style methods still leave researchers a large space to explore.

In this paper, our goal is to address the above issues and disadvantages of existing differentiable NAS algorithms, including the inefficiency of search algorithms, the enhancement in the structure selection process, and the handling of resource consumption. We propose a novel efficient NAS method named ZO-DARTS++, which aims to address these issues. The main contributions of this work are:
\begin{enumerate}
    \item We reformulate the optimization process of bi-level differentiable NAS problems. By incorporating the zeroth-order approximation method, the computationally prohibitive gradients of implicit variables can be easily substituted, making the search process accurate and efficient.

    \item We analyze the improper use of the probability normalization function for differentiable NAS methods. To increase interpretability, a novel sparsemax function is incorporated in ZO-DARTS++ to substitute the softmax function. Then we combine the temperature hyperparameter and annealing scheme with sparsemax, making the normalization results more sparse but still differentiable and solvable.

    \item We propose a size-variable search scheme that can effectively reduce resource consumption. Innovative inheritance and reuse of supernet results and parameters can reduce search costs while maintaining accuracy. The interplay with resource constraints can help ZO-DARTS++ design a more adaptable model for various platforms and applications.

    \item We showcase the effectiveness of the proposed resource-constrained size-variable NAS method. Comprehensive experiments on medical images characterized by diverse noise and under varying resource constraints validate the efficacy of ZO-DARTS++ in addressing the practical demands of fast and accurate medical image classification in resource-limited devices, even in the presence of noisy real-life data.
\end{enumerate}

The remainder of this paper is organized as follows. Section~\ref{sec2} provides a brief review of related work in the field. In Section~\ref{sec3}, we present a detailed description of our proposed ZO-DARTS++ algorithm. Section~\ref{sec4} provides experimental results demonstrating the effectiveness and efficiency of ZO-DARTS++ on medical imaging classification tasks. Finally, conclusions are drawn in Section~\ref{sec5}. 

\section{Related Work}\label{sec2}
In this section, we briefly review the development of differentiable NAS and its variants in resource-constrained scenarios. We also investigate the special model search scheme employed in such scenarios.

\subsection{Differentiable Neural Architecture Search}\label{sec2.1}
DARTS is a pioneering work on differentiable NAS frameworks, characterized by its continuous relaxation of the operation picking process. Although DARTS has been applied to various tasks, it still suffers from the unstable structure result.  Many studies~\cite{liang2019darts+, chu2020fair} have found that DARTS tends to select skip connections rather than operations with learnable parameters (e.g., convolution). Liang \emph{et al.}~\cite{liang2019darts+} conducts in-depth research and verifies this issue through a large number of experiments, then proposes a way to delete excessive skip connections by early-stopping. Besides, other tricks like operation-level dropout and fair operation-competing functions have been proposed. Meanwhile, the sampling scheme among operations in differentiable NAS also matters for better model performance. SNAS~\cite{xiesnas} presents a novel search gradient to reformulate the differentiable search process and reach higher performance. Moreover, Zhang \emph{et al.}~\cite{zhang2020data} introduces the Gumbel Softmax trick to convert architecture parameters to binary codes which reduces the estimation bias and is thus more accurate. In addition, there are also some works trying to improve the search efficiency by cutting the weak operation to shrink the search space~\cite{chen2021progressive} or sampling a subset of channels to reduce computation~\cite{xu2019pc}. 

The aforementioned works highlight that significant opportunities for further research remain in the field of differentiable NAS. Our previous work ZO-DARTS~\cite{xie2023zo}, laid the foundation by first incorporating zeroth-order approximation into DARTS-style differentiable NAS algorithm with the guidance of implicit gradient theorem~\cite{lorraine2020optimizing} and first-order optimality condition~\cite{tsaknakis2022implicit} to speed up the NAS process  but still presents first-tier performance on public datasets. Furthermore, we proposed ZO-DARTS+~\cite{xieefficient} to increase the interpretability and efficiency of the algorithm on real-world medical imaging datasets by introducing the sparsity-aware function. In this paper, we significantly extend the work by re-considering the computational ability constraint, proposing a novel size-variable search scheme, conducting more comprehensive experiments, and providing an advanced evaluation workflow for differentiable NAS methods.

\subsection{Resource-Constrained Neural Architecture Search Methods}\label{sec2.2}
Given the complexity of initial NAS techniques, novel trends concerning efficient search strategies and hardware-aware techniques are currently being developed to enhance the performance and accessibility of DL solutions. In addition to accuracy, computational overhead is another critical factor to consider. Hence, there is a growing trend for resource-aware NAS methods to balance between model performance and computation cost. POPNASv3~\cite{falanti2023popnasv3} utilizes a proxy function for model efficiency and builds a Pareto front for time and accuracy to guide the accuracy-harmless fast model selection. More works tend to add computational constraints like FLOPs, parameter size, and MACs to build multi-objective problems. For instance, CNAS~\cite{gambella_cnas_2022} takes parameter size, MACs, and activation size as constraints at the same time. RC-DARTS~\cite{jin2019rc} incorporates FLOPs and parameter size into its considerations, employing projection to manage the constraint function. Similarly, LC-NAS~\cite{li2022lc} implements efficiency control by training an additional latency regressor for various structures and applying constraints in a comparable manner. More recently, POMONAG was presented as a many-objective constrained algorithm belonging to the family of TransferableNAS dataset-aware techniques. Through its diffusion process called Many-Objective Reverse Diffusion Guidance, it is capable of generating extremely high-performance architectures that take multiple constraints into account simultaneously~\cite{lomurno2024pomonag}.

\begin{figure*}[htbp]
    \centering
    \includegraphics[width=0.9\linewidth]{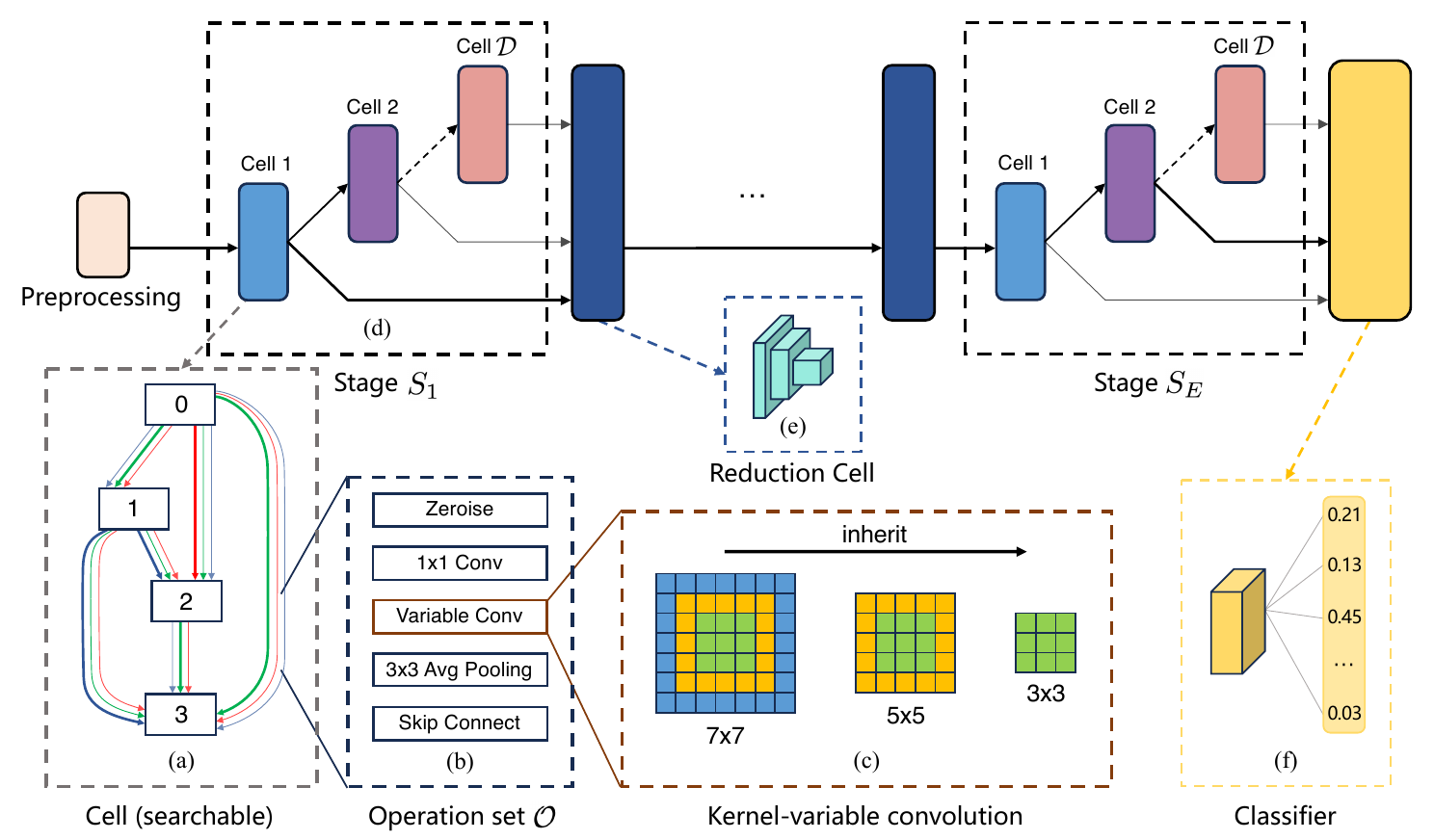}
    \caption{Overall search framework of ZO-DARTS++, which produces a flexible network model. The stages contain various cells, each has various operations, and the convolution kernel sizes and cell numbers can be adjusted during the search. (a) Basic searchable cell. (b) Operation set. (c) Kernel-variable convolution. (d) Depth-variable stage. (e) Reduction cell. (f) Final classifier.}
    \label{fig:pipeline}
\end{figure*}

\subsection{Flexible Neural Architecture Search Methods}
An adaptable NAS algorithm is essential when addressing constraints in real-world applications. While traditional EA- and RL-based methods are effective, they incur high search costs. To mitigate this, one-shot methods leverage weight sharing by training a supernet and deriving child architectures from it~\cite{2024ReCNAS}. The supernet, encompassing all candidate architectures, represents the entire search space as a single large network, with structure exploration integrated into its training process. Once-For-All~\cite{caionce}, a representative one-shot NAS algorithm, trains the supernet once using a progressive shrinking strategy, significantly reducing search costs by evaluating networks derived from the supernet. Building on the concepts of supernet and progressive shrinking, Lu \emph{et al.} introduced Neural Architecture Transfer, enabling the extraction of subnets tailored to different resource constraints~\cite{lu2021neural}. Similarly, Sarti \emph{et al.} expand these ideas with Neural Architecture Transfer 2, a paradigm designed to enhance efficiency in multi-objective NAS~\cite{sarti2023neural}. Although these flexible NAS approaches adapt well to model design under constraints, they remain rooted in EA-based methods, which still impose significant computational costs during the search process. However, the exploration of adaptable algorithms within the realm of differentiable NAS remains largely untapped, presenting a promising avenue for future research.

\section{Methodology}\label{sec3}
In this article, we propose ZO-DARTS++, a novel, flexible, and efficient search algorithm for the differentiable NAS problem, which is an effective extension of our previous work ZO-DARTS\cite{xie2023zo} and ZO-DARTS+\cite{xieefficient}. In Section~\ref{sec3.1}, we introduce the new bi-level problem formulation, which includes the constraint about the number of parameters, fundamental to designing smaller models. The search process is carried out by zeroth-order approximation to compute the gradient of loss function swiftly, which is detailed in Section~\ref{sec3.2}. In Section~\ref{sec3.3}, we present a novel integration of the sparsemax function with the temperature annealing scheme to generate architecture probabilities. This approach enhances the sparsity, plausibility, and interpretability of the output. The search process is further refined in Section~\ref{sec3.4} by incorporating size-variable convolution and adjustable model depth, effectively improving performance while reducing the overall model size. The complete pipeline of ZO-DARTS++ is illustrated in Fig.~\ref{fig:pipeline} and the algorithm Section~\ref{sec3.5}.

\subsection{Problem Definition}\label{sec3.1}
A differentiable NAS problem starts with the introduction of a fundamental building block, referred to as a cell in Fig.~\ref{fig:pipeline}(a). Searching for a cell that contains many operations and connecting edges has prevailed after the emergence of DARTS\cite{liu2018darts}. It is more economical than designing the whole model layer by layer. In this case, we can focus on a cell-level problem instead of a model-level problem. Each cell owns data inputs and hidden outputs denoted by numbered blocks. Edges between every $i^{th}$ and $j^{th}$ data block utilize a shared candidate set of operations $\mathcal{O}=\{o^{(i,j)}\}$ and each of them represents one different operation. Following the DARTS framework, we introduce architecture parameters $\balpha=\{\balpha^{(i,j)}\}$ to denote the importance weight of corresponding operations, which can transform the discrete NAS problem into a continuous optimization problem. Different colors and thicknesses of the edges in Fig.~\ref{fig:pipeline}(a) show the varying importance of operations. The mixed operation is defined by $\bar{o}^{(i,j)}(x)=\sum_{o\in\mathcal{O}}g(\balpha_o^{(i,j)})o^{(i,j)}(x)$, where $g(\cdot)$ denotes the function to normalize different architecture vectors $\balpha^{(i,j)}\in\mathbb{R}^{|\mathcal{O}|}$ as probability weights, and $o^{(i,j)}(x)$ indicates the specific output of the operation $o^{(i,j)}$ with input data $x$. Generally, this NAS problem can be expressed with two groups of parameters: the model parameters $\bw$ (for example, the weight of convolution filters) and the architecture parameters $\balpha$, which finally form the supernet. The optimization of parameters from one group is constrained by the other, which can be seen as an intertwined problem and formulated as a bi-level optimization problem:
\begin{equation}\label{eq1}
\begin{aligned}
&\min_{\balpha}\quad F(\balpha)=\Lval(\mathbf{w}^*(\balpha),\balpha)\\
&s.t.\quad\bw^*(\balpha)=\arg\min_{\bw}\Ltrain(\bw,\balpha).
\end{aligned}
\end{equation}
Here, $\Lval(\bw^*,\balpha)$ and $\Ltrain(\mathbf{w},\balpha)$ are loss functions on the validation and training datasets, corresponding to the upper-level (UL) and lower-level (LL) problems respectively. 

To shrink the search space, cells are grouped by different stages $S=\{S_1, S_2,..., S_E\}$. Previous methods~\cite{liu2018darts, liang2019darts+, he2020milenas} allow all cells to use the same set of $\balpha$, while we only allow cells to share the architecture parameters within each stage. Therefore, the final scale of the upper-level variable $\balpha=\{\balpha_{S_1}, \balpha_{S_2},..., \balpha_{S_E}\}$ is ${|S|\times|\mathcal{O}|\times N}$, where $N$ represents the number of edges within the cell. Different stages are connected by reduction cells with predefined ResBlocks, which can shrink the height and width of output feature maps but increase channel numbers. In addition, the structures of preprocessing and classifier modules are fixed to handle initial inputs and final outputs. 

To align with the trend of smart-device miniaturization, it is essential to account for the varying computational capabilities of different devices. Generally, computational abilities include the number of model parameters, latency, memory cost, etc. The parameter size serves as a direct indicator of the model's memory and computational overhead. In ZO-DARTS++, we incorporate mechanisms to effectively regulate model parameter size, thereby ensuring adaptability to resource-constrained environments. The expected number of parameters can be viewed as a mixed variable influenced by architecture parameters as well. For example, if we denote $\bc_o$ as the parameter number of operation $o\in\mathcal{O}$, then the expected number of parameters in one mixed edge $(i,j)$ is:
\begin{equation}\label{eq2}
    \bc^{(i,j)} = \sum_{o\in\mathcal{O}}g(\balpha_o^{(i,j)})\bc_o^{(i,j)}.
\end{equation}
Summing up the above cost of different edges from each cell forms the expected cost $C$. Hence, the final problem formulation considered by ZO-DARTS++ is the following:
\begin{equation}\label{eq3}
\begin{aligned}
\min_{\balpha}\quad F(\balpha)&=\Lval(\mathbf{w}^*(\balpha),\balpha)\\
s.t.\quad\bw^*(\balpha)&=\arg\min_{\bw}\Ltrain(\bw,\balpha),\\
C_L &\leq C\leq C_U.
\end{aligned}
\end{equation}
Here, $C_L$ and $C_U$ are the lower and upper bounds of the parameter number. In practice, only the upper bound is relevant when suppressing the size of the model. However, it is also necessary to keep a reasonable model scale, as excessively small models consist of insufficient learnable parameters, which can negatively impact accuracy. The above constraint is usually absorbed into the original problem as a penalty term. Thus, the whole problem can be expressed as follows:
\begin{equation}\label{eq4}
\begin{aligned}
    \min_{\balpha}\quad F(\balpha)&=\Lvalp(\mathbf{w}^*(\balpha),\balpha)\\
    s.t.\quad\bw^*(\balpha)&=\arg\min_{\bw}\Ltrain(\bw,\balpha),\\
    \Lvalp(\mathbf{w}^*(\balpha),\balpha)&=\Lval(\mathbf{w}^*(\balpha),\balpha)\\
    &\quad+\lambda_1\cdot\mathrm{max}(C-C_U, 0)\\
    &\quad+\lambda_2\cdot\mathrm{max}(C_L-C, 0).
\end{aligned}  
\end{equation}
We use non-negative hyperparameters $\lambda_1$ and $\lambda_2$ to control the level of penalization. Notice that we do not restrict the optimization process of model parameters $\bw$ in the LL problem, but only shape architecture parameters $\balpha$ in the UL problem. This penalty term can be applied via a ReLU function, which only activates when the number of parameters exceeds the budget. In this way, the search can be performed using gradient descent and solved smoothly.

\subsection{Zeroth-Order Approximation}\label{sec3.2}
To solve Eq. \eqref{eq4} accurately, the analytical gradient of $F(\balpha)$ can be derived by relying on the implicit function theorem~\cite{lorraine2020optimizing}:
\begin{equation}\label{eq5}
\nabla_{\!\balpha}F(\balpha)=\nabla_{\!\balpha}\Lvalp(\bw^*\!,\balpha)+\nabla_{\!\balpha}^T\!\bw^*\!(\balpha)\nabla_{\!\bw}\Lvalp(\bw^*\!(\balpha),\balpha),
\end{equation}
where
\begin{equation}\label{eq6}
    \nabla_{\!\balpha}\bw^*\!(\balpha)=-[\nabla_{\bw\bw}^2\Ltrain(\bw^{*},\balpha)]^{-1}\nabla_{\balpha\bw}^2\Ltrain(\bw^*,\balpha).
\end{equation}
Due to the impracticality of computing the Hessian as in Eq.~\eqref{eq6}, Xie \emph{et al.} propose ZO-DARTS~\cite{xie2023zo} to circumvent this obstacle. ZO-DARTS employs a zeroth-order approximation technique~\cite{nesterov2017random}:
\begin{equation}\label{eq7}
\tilde{\nabla}_{\balpha}F(\balpha)\triangleq\frac{F(\balpha+\mu\bu)-F(\balpha)}{\mu}\bu.
\end{equation}
As $\mu\rightarrow0$, the term $\frac{F(\balpha+\mu\bu)-F(\balpha)}{\mu}$ approximates the directional derivative $\nabla_{\balpha}^T F(\balpha)\bu$. Combining Eq.~\eqref{eq5} and Eq.~\eqref{eq7} we have:
\begin{equation}\label{eq8}
\scalebox{0.9}{$
\begin{aligned}
    \tilde{\nabla}_{\!\balpha}F(\balpha)\!
    &\approx\!\nabla_{\balpha}^T F(\balpha)\bu\bu\\
    &=\!\nabla_{\!\balpha}^T\Lvalp(\bw^*\!,\!\balpha)\bu\bu\!+\!\nabla_{\!\bw}^T\Lvalp(\bw^*\!(\balpha),\!\balpha)\nabla_{\!\balpha}\!\bw^*\!(\balpha)\bu\bu\\
    &=\!\nabla_{\!\balpha}^T\Lvalp(\bw^*\!,\!\balpha)\bu\bu\!+\![\nabla_{\!\balpha}\!\bw^*\!(\balpha)\bu]^T\nabla_{\!\bw}\Lvalp(\bw^*\!(\balpha),\!\balpha)\bu.
\end{aligned}
$}
\end{equation}
The term $\nabla_{\!\balpha}\bw^*\!(\balpha)\bu$ equals to $\frac{\bw^*(\balpha+\mu\bu)-\bw^*(\balpha)}{\mu}\bu$ according to the definition of directional derivative too, which can be calculated via an extra forward propagation. Meanwhile, the terms $\nabla_{\balpha}\Lvalp$ and $\nabla_{\bw}\Lvalp$ can be produced by back propagation. In addition, the additional gradient introduced by the penalty term is automatically included in $\nabla_{\balpha}\mathcal{L}_{val}$. Thus, $\tilde{\nabla}_{\balpha}F(\balpha)$ can be effectively approximated and computed swiftly.

\subsection{Sparsity-Aware Architecture Generation}\label{sec3.3}
The expression of indicator variables for different operations is a tricky problem. As introduced in Section~\ref{sec3.1}, a proper normalization function $g(\cdot)$ is necessary to formulate the NAS problem. Existing works consider this as a core problem. For example, ProxylessNAS~\cite{caiproxylessnas} employs a one-hot vector as a binary gate to represent the operation-selection process. Similarly, SNAS~\cite{xiesnas} adopts this scheme to sample different operations. In contrast, the approach introduced by DARTS uses continuous values to slack the operation-selection process. Traditionally, softmax is used to generate these continuous probability values from a set of architecture variables, which are then used to compute a weighted sum of operations. In DARTS-style methods, the mixed operation in edge $(i,j)$ is expressed as:
\begin{equation}\label{eq9}
\bar{o}^{(i,j)}(x)=\sum_{o\in\mathcal{O}}\frac{\exp(\balpha_o^{(i,j)})}{\sum_{o'\in\mathcal{O}}\exp(\balpha_{o'}^{(i,j)})}o(x).
\end{equation}
However, this method struggles with the complexity of bi-level NAS problems, as softmax tends to produce probabilities that do not diverge sufficiently and remain too similar across different operations~\cite{liang2019darts+}. This similarity is detrimental to selecting discrete operations, where a sparse probability distribution is preferable. To overcome these limitations, our approach employs the sparsemax function~\cite{martins2016softmax}, which projects the input vector $\balpha$ onto the probability simplex to produce a sparser probability distribution:
\begin{equation}\label{eq10}
    {\rm sparsemax}(\balpha):=\arg\min_{\bp\in\Delta^{K-1}}||\bp-\balpha||^2.
\end{equation}
The $(K-1)$-dimensional simplex, $\Delta^{K-1}:=\{\bp\in\mathbb{R}^K|\boldsymbol{1}^T\bp=1, \bp\geq0\}$, ensures that the sum of the probabilities is equal to one and each probability is non-negative, increasing the likelihood of obtaining a sparse result. This sparse expression of the probabilities is a properer normalization function to substitute softmax:
\begin{equation}\label{eq11}
    \bar{o}^{(i,j)}(x)=\sum_{o\in\mathcal{O}}{\rm sparsemax}_o(\balpha)o(x).
\end{equation}

Furthermore, sparsemax is designed to push outputs to extremes---close to 1 for selected operations and 0 for others---though it does not always eliminate non-zero elements. For softmax, a temperature parameter $\tau$ is commonly used to modulate the input values thus increasing sparsity:
\begin{equation}\label{eq12}
    \mathrm{softmax}'_o(\boldsymbol{\balpha}) = \frac{\exp(\balpha_o^{(i,j)}/\tau)}{\sum_{o'\in\mathcal{O}}\exp(\balpha_{o'}^{(i,j)}/\tau)}.
\end{equation}
As $\tau\rightarrow0$ this approach closely resembles an argmax function, ideal for achieving discrete selections. We effectively integrate this property into sparsemax by combining the trick of dividing input $\balpha_o^{(i,j)}$ by $\tau$. To further enhance this property, we propose an annealing strategy to reduce $\tau$ by a manually chosen factor $a\leq 1$ every $m$ epochs, dividing the system towards a one-hot vector. Therefore, the final normalization function comes to:
\begin{equation}\label{eq13}
    \mathrm{sparsemax}'(\balpha)=\arg\min_{\bp\in\Delta^{k-1}}||\bp - \balpha/(\tau * a^{n//m})||^2,
\end{equation}
where $n$ is the total number of epochs, and $//$ represents the integer division. By adjusting the factors $a$ and $m$, the algorithm efficiently accelerates the search process while promoting sparsity, closely matching the need for fast and automatic model generation.

\subsection{Size-Variable Search Scheme}\label{sec3.4}
To reduce the search cost whilst chasing the most suitable model structure for platforms with different capacities, we draw inspiration from the Once-For-All method~\cite{caionce} and creatively propose a size-variable search scheme concerning both the kernel size and depth of the DL model for the differentiable NAS algorithm. 

For an ordinary convolution from the operation set $\mathcal{O}$, its kernel size is usually specified in advance. In differentiable NAS, a general way to produce smaller models is to declare convolutions with various kernel sizes $k\in\mathcal{K}$ and treat them as different operations, which increases both memory and time cost during the search process. To address this, we consider a kernel-variable convolution search scheme that activates only the largest kernel at the beginning. After thorough exploration, the algorithm then switches the single-kernel convolution to one mixed convolution, which includes many convolutions with different kernel sizes and probability weights indicating importance. Denote $\bbeta$ as the probability weight vector for different sizes of convolutions, then the mixed convolution in edge $(i,j)$ is expressed as:
\begin{equation}\label{eq14}
    \overline{conv}^{(i,j)}(x)=\sum_{k\in\mathcal{K}}\mathrm{sparsemax}'_k(\bbeta)conv_k^{(i,j)}(x).
\end{equation}
We also utilize sparsemax to preserve sparsity better when choosing different sizes of kernels. Notably, smaller kernels will inherit the center part of the largest kernel. This process is illustrated in Fig.~\ref{fig:pipeline}(c). When the kernel-variable search is activated, the kernel parameters and their corresponding probability weights for different kernels can jointly form a linear combination of the kernel parameters from the largest original kernel. The only difference exists in the new bias that follows each size of the kernel. From this angle, our scheme swiftly turns one convolution into three parallel convolutions and avoids imposing difficulties on the optimization process. This reduces the algorithm cost to a reasonable level, making it easier to search compared to inspecting different kernels from scratch. 

The depth of the model is another adjustable hyperparameter. Generally, DL models require numerous layers, and determining the optimal quantity is challenging. In the field of NAS, algorithms either predefine the number of layers or progressively increase network depth. For cell-based models, layers are replaced by cells, which are organized into multiple stages in sequence. The number of cells within each stage is more manageable, with deeper stages representing more comprehensive processing. Here we  develop a novel depth-variable scheme inspired by the early-exit trick. Let $\bgamma$ be the probability weight vector for the outputs of different cells with depth $d\in\mathcal{D}$ from one stage. Probabilities corresponding to different lines in Fig.~\ref{fig:pipeline}(d) represent diverse importance, and the mixed output of the $e^{th}$ stage has the following form:
\begin{equation}\label{eq15}
\overline{output}^e=\sum_{d\in\mathcal{D}}\mathrm{sparsemax}'_d(\bgamma)output_d^e.
\end{equation}
The design logic of cell-based models originates from ResNet which stacks layers and uses skip connections to help build the net. In this case, our mixed output of each stage can be regarded as cells with different levels of skip connections, which is better in search compared to the single output from the deepest cell. Similarly, the algorithm does search on the deepest model in the initial epochs, thus the output of each stage is simply the output of the last cell. Once the exploration threshold is reached, the output of each stage is replaced with the mixed output from different cells. When the search process is finished, outputs with larger probability weights are more likely to be picked. Thus, the most suitable cell number is obtained.

In addition, the size-variable search scheme maintains the validity of the previously established parameter computation method. For one mixed convolution $\overline{conv}^{(i,j)}$, the parameter count is calculated as:
\begin{equation}\label{eq16}
\bc_{\overline{conv}}^{(i,j)}=\sum_{k\in\mathcal{K}}\mathrm{sparsemax}'_k(\bbeta)\bc_{conv_k}^{(i,j)}.
\end{equation}
Here, $\mathrm{sparsemax}'(\cdot)$ acts as the probability weight normalization function $g(\cdot)$ in Eq.~\eqref{eq2}. The total number of parmeters within a stage $S_e$ is then computed by weighting the contributions from all candidate cells too:
\begin{equation}
\bc^e=\sum_{d\in\mathcal{D}}\mathrm{sparsemax}'_d(\bgamma)\bc_{d}^e,
\end{equation}
where $\bc_d^e$ refers to the consumption of the $d^{th}$ cell in stage $S_e$. The above two equation corresponds to Eq.~\eqref{eq14} and Eq.~\eqref{eq15} respectively. This approach ensures that the expected parameter count reflects the probabilistic architecture design while accommodating the size-variable search.

\subsection{Algorithm}\label{sec3.5}
The entire search process implemented by ZO-DARTS++ is summarized in Algorithm~\ref{alg1}. The model will be searched for $n$ epochs, and the entire process will be divided into two segments defined by introducing a threshold $\theta$. In the beginning, the algorithm will only search for the largest model (i.e., the largest cell number and the widest kernel). When the threshold is reached, the architecture parameters for smaller structures will be activated in line 5. The core iteration part of one epoch is in lines 7-18. The algorithm generates random variables $\mu$ and $\bu$ needed for zeroth-order approximation in lines 8. A surrogate model is also introduced in line 9 to calculate the zeroth-order difference. After that, the algorithm updates model parameters $\bw$ for $T$ rounds in lines 10-14, corresponding to the optimization process of the LL problem. This is to ensure the relatively proper $\bw$ for the update in the UL problem. Next, the architecture parameters $\balpha'$ will be updated from lines 14-16, according to Eq.~\eqref{eq8}. Finally, a supernet containing all the information is saved for evaluation.
\begin{algorithm}[htbp]
    \renewcommand{\algorithmicrequire}{\textbf{Initialize:}}
    \renewcommand{\algorithmicensure}{\textbf{Output:}}
    \caption{ZO-DARTS++}  \label{alg1}
    \begin{algorithmic}[1]
    \REQUIRE Operation set $\mathcal{O}$, probability weight $\balpha$, kernel and exit probability weight $\bbeta$ and $\bgamma$, dataset $D$;
    \ENSURE Supernet;
    \WHILE{$epoch < n$}
    \IF{$epoch < \theta$}
    \STATE Set $\balpha' = \balpha$, $\Lvalp=\Lval$
    \ELSE{}
    \STATE Set $\balpha' = \{\balpha, \bbeta, \bgamma\}$, $\Lvalp=\Lval+\lambda_1\cdot\mathrm{max}(C-C_U, 0)+\lambda_2\cdot\mathrm{max}(C_L-C, 0)$
    \ENDIF
    \FOR{$D_{train}$, $D_{val}$ in $D$}
    \STATE Set $\mu$ and generate random vector $\bu$ with $||\bu||^2=1$;
    \STATE Set $\tilde{\balpha}'=\balpha'+\mu\bu$, $\tilde{\bw}=\bw$, and $t=1$;
    \WHILE{$t < T$}
    \STATE Update $\bw=\bw-\eta_{\bw}*\nabla_{\bw}\Ltrain(\bw, \balpha')$;
    \STATE Update $\tilde{\bw}=\tilde{\bw}-\eta_{\bw}*\nabla_{\tilde{\bw}}\Ltrain(\tilde{\bw}, \tilde{\balpha}')$;
    \STATE $t = t + 1$;
    \ENDWHILE, Let $\bw^*\!=\!\bw$;
    \STATE Compute $\nabla_{\balpha'}\bw^*(\balpha')\bu=\frac{\tilde{\bw}(\tilde{\balpha}')-\bw(\balpha')}{\mu}$;
    \STATE Compute $\tilde{\nabla}_{\balpha'}F(\balpha')
    =\nabla_{\!\balpha'}^T\Lvalp(\bw^*\!,\!\balpha')\bu\bu$\\
    $\qquad\qquad+[\nabla_{\!\balpha'}\bw^*\!(\balpha')\bu]^T\nabla_{\!\bw}\Lvalp(\bw^*\!(\balpha'),\balpha')\bu$;
    \STATE Update $\balpha'=\balpha'-\eta_{\balpha'}*\tilde{\nabla}_{\balpha'}F(\balpha')$;
    \ENDFOR
    \ENDWHILE
    \RETURN Supernet
    \end{algorithmic}
\end{algorithm}
\section{Experimental Analysis}\label{sec4}
In this section, we conduct a quantitative analysis to assess the effectiveness and efficiency of ZO-DARTS++\footnote{https://github.com/HikariX/ZO-DARTS\_PP}. We begin by detailing the experimental setup with specialized medical imaging datasets and emphasizing our novel approach to deriving constraints for the search process. To evaluate the performance, we perform extensive experiments across a diverse range of datasets. Furthermore, we provide an in-depth analysis of our method performance under resource constraints. The discussion on reliability and interpretability is then conducted to highlight ZO-DARTS++ advantages. Additionally, a comprehensive comparison with a state-of-the-art evolutionary algorithm is also provided.

\subsection{Dataset and Settings}
To assess the effectiveness and efficiency of our proposed approach, we conduct experiments on 10 medical imaging datasets from the MedMNIST classification tasks: PathMNIST, BloodMNIST, DermaMNIST, PneumoniaMNIST, TissueMNIST, OCTMNIST, BreastMNIST, OrganAMNIST, OrganCMNIST, and OrganSMNIST\cite{medmnistv2}. Facing the potential computational capacity limit of hospital clusters, advancing research on fast, accurate, and flexible NAS methods tailored to such data is of critical importance. These datasets are derived from real-world clinical settings and may include noise artifacts introduced during the acquisition process by medical imaging devices, making NAS more challenging. The first three datasets comprise RGB images with three channels, whereas the remaining ones consist of single-channel grayscale images. The dataset sizes range from 780 to 236,386 samples, reflecting a broad spectrum of data complexity. During the searching process, the training and validation datasets are evenly split to address the bi-level optimization problem, following the same approach as DARTS. In the retraining phase, the original dataset test-split ratio is maintained. In addition, we set the batch sizes of different datasets according to their scale and will adjust a little if they are hard to get good model performance. Details are provided in Table~\ref{tab:dataset}.
\begin{table}[htbp]
    \centering
    \setlength{\tabcolsep}{2pt}
    \footnotesize
    \caption{Dataset information.}
    \label{tab:dataset}
    \begin{tabular}{lccc}
    \toprule
        Dataset & \# Training/Validation/Test & Search batch & Retrain batch\\
        \midrule
        PathMNIST & 89,996/10,004/7,180 & 512 & 128\\
        BloodMNIST & 11,959/1,712/3,421 & 128 & 32\\
        DermaMNIST & 7,007/1,003/2,005 & 128 & 32\\
        PneumoniaMNIST & 4,708/524/624 & 32 & 16\\
        TissueMNIST & 165,466/23,640/47,280 & 512 & 256\\
        OCTMNIST & 97,477/10,832/1,000 & 512 & 128\\
        BreastMNIST & 546/78/156 & 16 & 64\\
        OrganAMNIST & 34,561/6,491/17,778 & 256 &128\\
        OrganCMNIST & 12,975/2,392/8,216 & 256 & 128\\
        OrganSMNIST & 13,932/2,452/8,827 & 256 & 128\\
    \bottomrule
    \end{tabular}
\end{table}

The baselines for comparison we consider include ResNet-18\cite{he2016deep} as a standard for manually designed models and included AutoKeras\cite{JMLR:v24:20-1355} and Google AutoML Vision\cite{google_automl} for NAS models. In addition, we evaluate four differentiable NAS methods: DARTS~\cite{liu2018darts}, MileNAS~\cite{he2020milenas}, accompanied with our two prior works ZO-DARTS~\cite{xie2023zo} and ZO-DARTS+~\cite{xieefficient}. They all operate within a search space defined by NAS-Bench-201~\cite{dong2020bench} with five operations: Zeroise, Skip Connect, 1x1 Conv, 3x3 Conv, and 3x3 Avg Pooling. 
Notice that ZO-DARTS only includes the optimization process mentioned in Section~\ref{sec3.2}, and ZO-DARTS+ is equipped with sparsity-aware function in Section~\ref{sec3.3}. Additionally, differentiable algorithms in the experiment follow the same hyperparameter setting listed above. We execute each search algorithm three times on different seeds and consider the mean and variance of the performance metrics obtained.  Finally, we also introduce in Section~\ref{sec:popnas} a comparison with POPNASv3, as representative of the population-based methods.

For what concerns hyperparameters, the model in our ZO-DARTS++ framework includes $|S|=3$ stages, each containing $N=3$ cells initially. We perform $n=50$ search epochs, and the threshold epoch is set to $\theta=20$ for size-variable search and constraints. ZO-DARTS++ updates the architecture parameters $\balpha$ only after the model parameters $\bw$ are optimized $T=10$ times. The infinitesimal which controls the zeroth-order gradient is $\mu=0.005 * |\balpha'|$. An initial temperature $\tau=1.5$ is used to encourage exploration in the early stages of the search. The annealing factor is set to $a=0.75$, with an interval of $m=5$, allowing a gradual refinement of the model configurations. Regarding the control of constraints, we set the penalty factors $\lambda_1$ and $\lambda_2$ both equal to $15/(\tau*a^{n//m})$, an empirical factor linked to annealing temperature. This will allow the algorithm to be loose at the beginning and comply with constraints in the later epochs of search.

The experiments in the search phase are executed using PyTorch 3.8.10 and CUDA 11.8 on one NVIDIA GeForce RTX 4090 and Intel Xeon Silver 4314 Processor.

\subsection{Advanced Evaluation Workflow}\label{sec:evaluation}
Differentiable NAS algorithms evaluate their result using a special retraining strategy. Typically, the final model structures are composed of operations corresponding to the highest probabilities. It is somehow reasonable but received criticism~\cite{liang2019darts+, chu2020fair} due to the lack of information and the structural gap between the mixed supernet and the discrete structure. Unlike traditional methods, ZO-DARTS++ will derive the final discrete architecture from the learned architecture probabilities of the supernet, ensuring that the most promising operations are selected while preserving the probabilistic nature of the search. For instance, an operation with a higher probability among an edge candidate operations is more likely to be chosen. In this way, a sparser supernet would ensure a more stable sampling result, which conforms to our sparsity-aware search scheme. The final model accuracy is then determined by retraining a structure to full convergence. 

The search algorithm will be executed under three different seeds. To better make use of the searched supernet information, we do the sampling process three times for each searched supernet. Conducting multiple experiments enhances the reliability of our findings by reducing the impact of random variations and minimizing biases or measurement errors. Additionally, it increases the robustness and generalizability of our conclusions while offering a broader basis for meaningful insights. This strategy is also applied to other differentiable NAS algorithms in the comparison. Beyond that, we will omit models that did not meet the size constraint when the size control is activated. In addition, we would like to focus on the performance of the search algorithm itself rather than the endless fine-tuning in the retraining process. During retraining, structures will be discarded at the $20^{th}$ epoch once their accuracy falls below 30\% or if the improvement compared to the initial epoch is less than 1\%.

\subsection{Model Size Constraints Determination}
\begin{figure*}[htbp]
    \centering
    \includegraphics[width=1\linewidth]{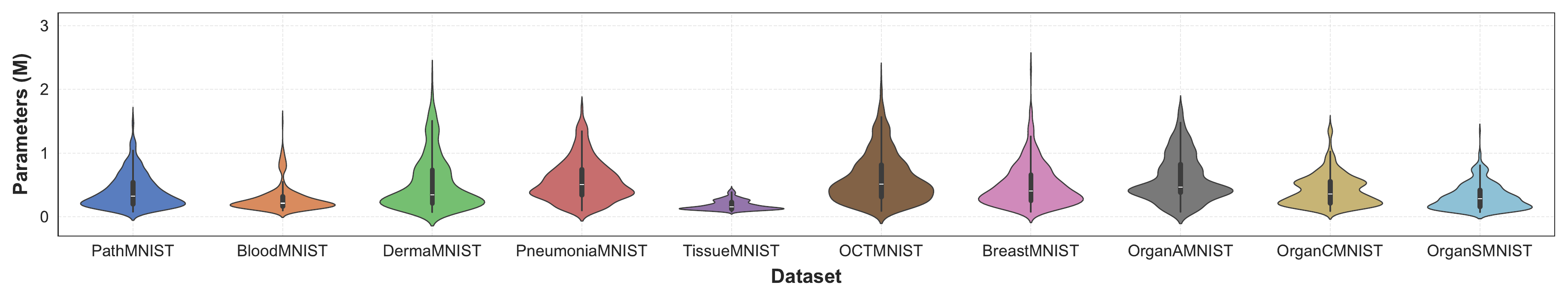}
    \caption{Violin plots of parameter numbers (million) distribution of models across various datasets. Each violin represents the range, density, and median parameter size for models sampled from supernets searched without constraints on the respective datasets.}
    \label{fig:size}
\end{figure*}

The vast range of datasets indicates the possibility of various model sizes. To make the constrained model more reasonable, we execute the search process by launching the unconstrained ZO-DARTS++ algorithm and getting different supernets under three seeds for each dataset first. Next, 300 structures from each supernet are sampled separately using the same method in Section~\ref{sec:evaluation}. Multiple results will improve reliability, reduce errors, ensure robustness, and support rigorous statistical analysis. The model size distribution of nearly a thousand structures is shown in Fig.~\ref{fig:size}. The violin plot of all datasets exhibits a similar preference of structures for being small as the majority and has a long tail for super-large models. The thicker part reflects the higher density of the structures with a specific parameter size. The internal box plot represents the lower, median, and upper quartiles, which illustrate the distribution characteristics of the model sizes within the dataset. For example, OCTMNIST has model candidates larger than those of other datasets, while the majority of models on TissueMNIST remain under 0.5M in parameter size. This shows the underlying taste of the algorithm for different model structures and sizes across various datasets.

Guided by these distributions, we derive constraints using different percentiles. Three intervals $[0, 20\%]$, $[40\%, 60\%]$, and $[80\%, 95\%]$, are defined as ``S", ``M", and ``L", respectively, representing different lower and upper bounds for model sizes. It is worth noting that we use $0$ instead of $0\%$ as a lower bound since these models inherently have nonzero minimum sizes. Additionally, to mitigate the impact of outliers, the percentile $95\%$ is set as the maximum upper bound for model sizes instead of $100\%$. This decision is justified by the observed long-tail distribution of model sizes.

\begin{table*}[htbp]
    \centering
    \setlength{\tabcolsep}{2pt}
    \footnotesize
    \caption{Accuracy comparison (\%) with baselines and state-of-the-art algorithms on the MedMNIST dataset.}
    \label{tab:acc}
    \scalebox{0.85}{
    \begin{tabular}{lccccccccccc}
        \toprule
        Dataset & PathMNIST & BloodMNIST & DermaMNIST & PneumoniaMNIST & TissueMNIST & OCTMNIST &  BreastMNIST &  OrganAMNIST & OrganCMNIST &  OrganSMNIST & Avg\\ 
        \midrule
        ResNet-18            & \textbf{90.7} & 95.8 & 73.5 & 85.4 & \underline{67.6} &  74.3 &  86.3 &  \underline{93.5} & 90.0 &  78.2 & \underline{83.53} \\ 
        AutoKeras           & \underline{83.4} & \underline{96.1} & 74.9 & 87.8 & \textbf{70.3} &  76.3 &  83.1 &  90.5 & 87.9 &  \textbf{81.3} & 83.16 \\ 
        Google AutoML       & 72.8 & \textbf{96.6} & \underline{76.8} & \textbf{94.6} & 67.3 &  77.1 &  86.1 &  88.6 & 87.7 &  74.9 & 82.25 \\ 
        \midrule
        DARTS               & 78.6$\pm$3.50 & 86.7$\pm$3.74 & 75.2$\pm$0.92 & 93.7$\pm$0.64 & 63.7$\pm$1.40 &  79.1$\pm$1.73 &  87.0$\pm$1.01 &  92.8$\pm$0.86 & 89.1$\pm$1.35 &  76.1$\pm$2.72 & 82.20 \\
        MiLeNAS               & 81.1$\pm$4.07 & 86.8$\pm$3.76 & 75.6$\pm$0.73 & 94.0$\pm$0.58 & 64.2$\pm$1.34 &  78.1$\pm$3.34 &  87.5$\pm$1.33 & 93.1$\pm$0.85 & 89.8$\pm$1.37 &  76.7$\pm$2.07 & 82.69 \\
        POPNASv3    & 72.6$\pm$3.27 & 93.5$\pm$0.75 & \textbf{77.9}$\pm$0.15 & 93.9$\pm$0.80 & 64.7$\pm$0.94 & \textbf{84.7}$\pm$1.31 & 87.4$\pm$3.03 & 88.4$\pm$1.00 & 86.6$\pm$2.38 & 74.3$\pm$1.13 & 82.40\\
        ZO-DARTS            & 81.6$\pm$5.39 & 85.6$\pm$5.82 & 75.8$\pm$0.71 & 94.3$\pm$0.49 & 64.9$\pm$1.12 &  79.8$\pm$1.81 & 86.1$\pm$5.12 & 93.1$\pm$1.32 & 89.6$\pm$1.89 & 77.6$\pm$1.58 & 82.84 \\
        \midrule
        ZO-DARTS+    & 78.9$\pm$4.75 & 88.3$\pm$6.55 & 76.2$\pm$0.82 & 94.1$\pm$0.69 & 63.7$\pm$1.85 &  \underline{80.1}$\pm$1.74 & 86.3$\pm$5.13 & 93.1$\pm$0.76 & 90.1$\pm$1.22 & 77.2$\pm$1.43 & 82.80 \\
        ZO-DARTS++    & 80.7$\pm$2.83 & 89.2$\pm$1.99 & 76.3$\pm$0.80 & 94.0$\pm$0.69 & 62.9$\pm$2.19 & 79.4$\pm$1.20 & 87.1$\pm$5.44 & 93.3$\pm$1.02 & \underline{90.5}$\pm$0.75 & 77.4$\pm$1.51 & 83.08 \\
        ZO-DARTS++ S    & 80.6$\pm$3.30 & 89.2$\pm$2.84 & 76.1$\pm$0.98 & 94.3$\pm$0.48 & 62.8$\pm$0.85 &  76.7$\pm$2.69 & \underline{88.5}$\pm$1.71 & 92.9$\pm$0.40 & 89.6$\pm$0.52 & 76.9$\pm$0.76 & 82.76 \\
        ZO-DARTS++ M    & 80.8$\pm$4.59 & 89.1$\pm$2.46 & 75.3$\pm$3.20 & \underline{94.4}$\pm$0.48 & 64.3$\pm$1.40 &  80.0$\pm$1.38 & \underline{88.5}$\pm$1.16 & 93.2$\pm$0.74 & 90.4$\pm$0.58 & 77.8$\pm$0.61 & 83.38 \\
        ZO-DARTS++ L    & 80.4$\pm$2.41 & 91.2$\pm$1.37 & 76.0$\pm$0.64 & 94.1$\pm$0.70 & 65.6$\pm$1.15 &  79.9$\pm$1.03 & \textbf{89.1}$\pm$1.24 & \textbf{94.2}$\pm$0.57 & \textbf{90.9}$\pm$0.83 & \underline{78.6}$\pm$0.28 & \textbf{84.00} \\
        \bottomrule
    \end{tabular}
}
\end{table*}

\begin{table*}[t]
    \centering
    \setlength{\tabcolsep}{2pt}
    \footnotesize
    \caption{Search Time (second) comparison with baselines and state-of-the-art algorithms on the medmnist dataset.}
    \label{tab:time}
    \scalebox{0.85}{
    \begin{tabular}{lccccccccccc}
        \toprule
        Dataset & PathMNIST & BloodMNIST & DermaMNIST & PneumoniaMNIST & TissueMNIST & OCTMNIST &  BreastMNIST &  OrganAMNIST & OrganCMNIST & OrganSMNIST & Avg \\ 
        \midrule
        DARTS               & 3470.6 & 1168.0 & 694.8 & 1796.1 & 6501.6 & 3858.4 & 436.5 & 1881.4 & 701.6 & 755.0 & 2126.40 \\
        MiLeNAS               & 2902.6 & 946.8 & 576.3 & 1407.8 & 5426.0 & 3229.3 & 352.4 & 1469.3 & 552.3 & 598.7 & 1746.15 \\
        POPNASv3    & 26833.0 & 26742.8 & 15854.4 & 25937.6 & 96066.0 & 59038.4 & 9247.0 & 26955.9 & 12724.3 & 13439.3 & 31283.87\\
        ZO-DARTS            & 1589.8 & 503.3 & 298.2 & 736.9 & 2982.4 & 1731.9 & 184.1 & 735.2 & 290.1 & 315.5 & 936.74 \\
        \midrule
        ZO-DARTS+    & \textbf{1279.6} & \textbf{400.4} & \textbf{247.3} & \textbf{606.6} & \textbf{2405.0} & \textbf{1381.2} & \textbf{148.3} & \textbf{617.2} & \textbf{239.8} & \textbf{252.6} & \textbf{757.80}\\ 
        ZO-DARTS++    & 2670.6 & 576.4 & 346.1 & 1009.6 & 5124.9 & 2908.7 & 212.3 & 1121.7 & 423.0 & 448.2 & 1484.15 \\
        \bottomrule
    \end{tabular}
    }
\end{table*}

\subsection{ZO-DARTS++ Performance}
Table~\ref{tab:acc} shows the optimal performance achieved by each algorithm on different datasets and their average, with baseline data taken directly from the MedMNIST homepage\footnote{https://medmnist.com}. ZO-DARTS++ is our algorithm without any constraints, while ZO-DARTS+ and ZO-DARTS are our two previous works without size-variable search and even without sparsity-aware scheme. ZO-DARTS++ outperforms ResNet-18 on five datasets and wins both AutoKeras and Google AutoML on six datasets. For the remaining datasets, it demonstrates comparable performance with minimal gaps. ZO-DARTS++ also wins other popular methods on most datasets or just keeps a tiny distance, demonstrating robust performance across the board. It even surpasses the original DARTS on nine datasets. The average accuracy across different datasets also supports our conclusion. 

We attribute the effectiveness of ZO-DARTS++ to its advanced structural design. Traditional DARTS-style methods enforce a uniform structure across all cells, oversimplifying the search space and limiting optimization flexibility. In contrast, ZO-DARTS++ expands the architecture parameter space, making it three times larger than that of conventional methods, thereby enabling more diverse structural combinations. While cells within the same stage share a common topology, ZO-DARTS++ allows different topologies across stages, offering greater design flexibility and adaptability compared to traditional approaches. Furthermore, the size-variable search scheme somehow reasonably extends the search space, enabling the model to have different kernel sizes and cell numbers. In return, the increased model complexity of ZO-DARTS++ provides an overall performance increase from 82.80\% to 83.08\% compared to ZO-DARTS+.

ZO-DARTS++ not only wins in model accuracy but also achieves a significant reduction in search time, as detailed in Table~\ref{tab:time}. We examined the convergence graph and found that algorithms with the sparsity-aware scheme have divergent probabilities. It was first verified in ZO-DARTS+~\cite{xieefficient}. Given the rapid convergence, we do early-stopping around the $40^{th}$ epoch of search and record the search time of ZO-DARTS++ (and ZO-DARTS+) while other methods take the result from the $50^{th}$ epoch. This phenomenon is further discussed in Section~\ref{sec:reliability}. To seek better and more complex models, ZO-DARTS++ inevitably needs a longer search time than its two predecessors. But still, ZO-DARTS++ saved 38.6\% of the time on average during the search compared to the original DARTS. Our method can reach a good equilibrium of performance and efficiency. The efficiency underlines the suitability of ZO-DARTS++ for the development of real-life applications, especially medical image classification models. 
\begin{table*}[t]
    \centering
    \setlength{\tabcolsep}{2pt}
    \footnotesize
    \caption{Parameter numbers (million) comparison with baselines and state-of-the-art algorithms on the MedMNIST dataset.}
    \label{tab:param}
    \scalebox{0.85}{
    \begin{tabular}{lccccccccccc}
        \toprule
        Dataset & PathMNIST & BloodMNIST & DermaMNIST & PneumoniaMNIST & TissueMNIST & OCTMNIST &  BreastMNIST &  OrganAMNIST & OrganCMNIST & OrganSMNIST & Avg \\ 
        \midrule
        DARTS            & 0.21$\pm$0.1 & 0.25$\pm$0.1 & 0.21$\pm$0.1 & 0.31$\pm$0.1 & \underline{0.16}$\pm$0.1 & \textbf{0.15}$\pm$0.1 & 0.22$\pm$0.1 & \underline{0.22}$\pm$0.1 & 0.25$\pm$0.1 & \underline{0.22}$\pm$0.1 & \underline{0.220} \\
        MiLeNAS            & 0.30$\pm$0.1 & 0.21$\pm$0.2 & 0.29$\pm$0.1 & 0.32$\pm$0.2 & 0.28$\pm$0.1 & 0.33$\pm$0.2 & 0.33$\pm$0.1 & 0.30$\pm$0.1 & \underline{0.21}$\pm$0.1 & 0.24$\pm$0.2 & 0.281 \\
        POPNASv3            & \textbf{0.08}$\pm$0.1 & \underline{0.20}$\pm$0.2 & \underline{0.17}$\pm$0.1 & \underline{0.26}$\pm$0.1 & 0.52$\pm$0.0 & 0.33$\pm$0.1 & \textbf{0.05}$\pm$0.0 & 0.39$\pm$0.2 & 0.39$\pm$0.3 & 0.26$\pm$0.1 & 0.265 \\
        ZO-DARTS            & 0.23$\pm$0.1 & 0.24$\pm$0.1 & 0.32$\pm$0.2 & 0.28$\pm$0.1 & 0.31$\pm$0.1 & 0.30$\pm$0.2 & 0.32$\pm$0.2 & 0.30$\pm$0.1 & 0.33$\pm$0.2 & 0.27$\pm$0.1 & 0.290 \\
        \midrule
        ZO-DARTS+            & 0.18$\pm$0.1 & 0.29$\pm$0.1 & 0.21$\pm$0.1 & 0.43$\pm$0.1 & 0.30$\pm$0.3 & 0.38$\pm$0.1 & 0.35$\pm$0.1 & \underline{0.22}$\pm$0.1 & 0.28$\pm$0.1 & 0.32$\pm$0.1 & 0.296 \\ 
        ZO-DARTS++            & 0.45$\pm$0.4 & 0.24$\pm$0.1 & 0.56$\pm$0.7 & 0.65$\pm$0.3 & 0.17$\pm$0.0 & 0.59$\pm$0.2 & 0.55$\pm$0.3 & 0.75$\pm$0.6 & 0.31$\pm$0.2 & 0.27$\pm$0.2 & 0.454 \\
        ZO-DARTS++ S    & \underline{0.11}$\pm$0.0 & \textbf{0.12}$\pm$0.0 & \textbf{0.14}$\pm$0.0 & \textbf{0.18}$\pm$0.1 & \textbf{0.09}$\pm$0.0 & \underline{0.16}$\pm$0.0 & \underline{0.15}$\pm$0.0 & \textbf{0.21}$\pm$0.1 & \textbf{0.11}$\pm$0.0 & \textbf{0.10}$\pm$0.0 & \textbf{0.137} \\
        ZO-DARTS++ M    & 0.34$\pm$0.0 & 0.21$\pm$0.0 & 0.37$\pm$0.1 & 0.52$\pm$0.1 & \underline{0.16}$\pm$0.0 & 0.53$\pm$0.1 & 0.41$\pm$0.0 & 0.48$\pm$0.1 & 0.40$\pm$0.0 & 0.27$\pm$0.0 & 0.369 \\
        ZO-DARTS++ L    & 0.82$\pm$0.1 & 0.48$\pm$0.1 & 1.18$\pm$0.2 & 1.05$\pm$0.2 & 0.27$\pm$0.0 & 1.21$\pm$0.2 & 0.87$\pm$0.1 & 1.12$\pm$0.1 & 0.71$\pm$0.1 & 0.62$\pm$0.1 & 0.833 \\
        \bottomrule
    \end{tabular}
    }
\end{table*}

\begin{figure*}[htbp]
    \centering
    \includegraphics[width=1\linewidth]{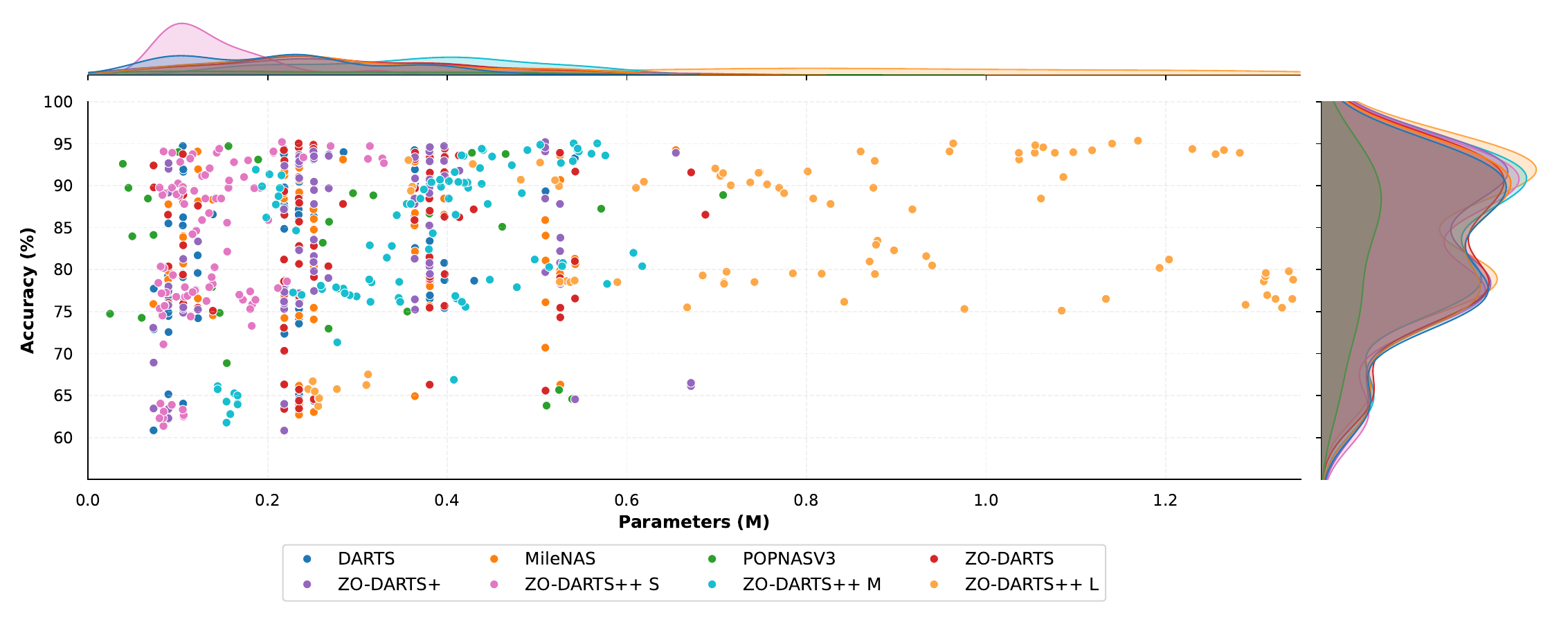}
    \caption{Joint distribution of model accuracy (\%) and parameter numbers (million) for different NAS methods. Each point represents a specific model structure, with density distributions of accuracy (top) and parameter size (right) shown as marginal plots.}
    \label{fig:density}
\end{figure*}

\begin{figure*}[htbp]
    \centering
    \includegraphics[width=1\linewidth]{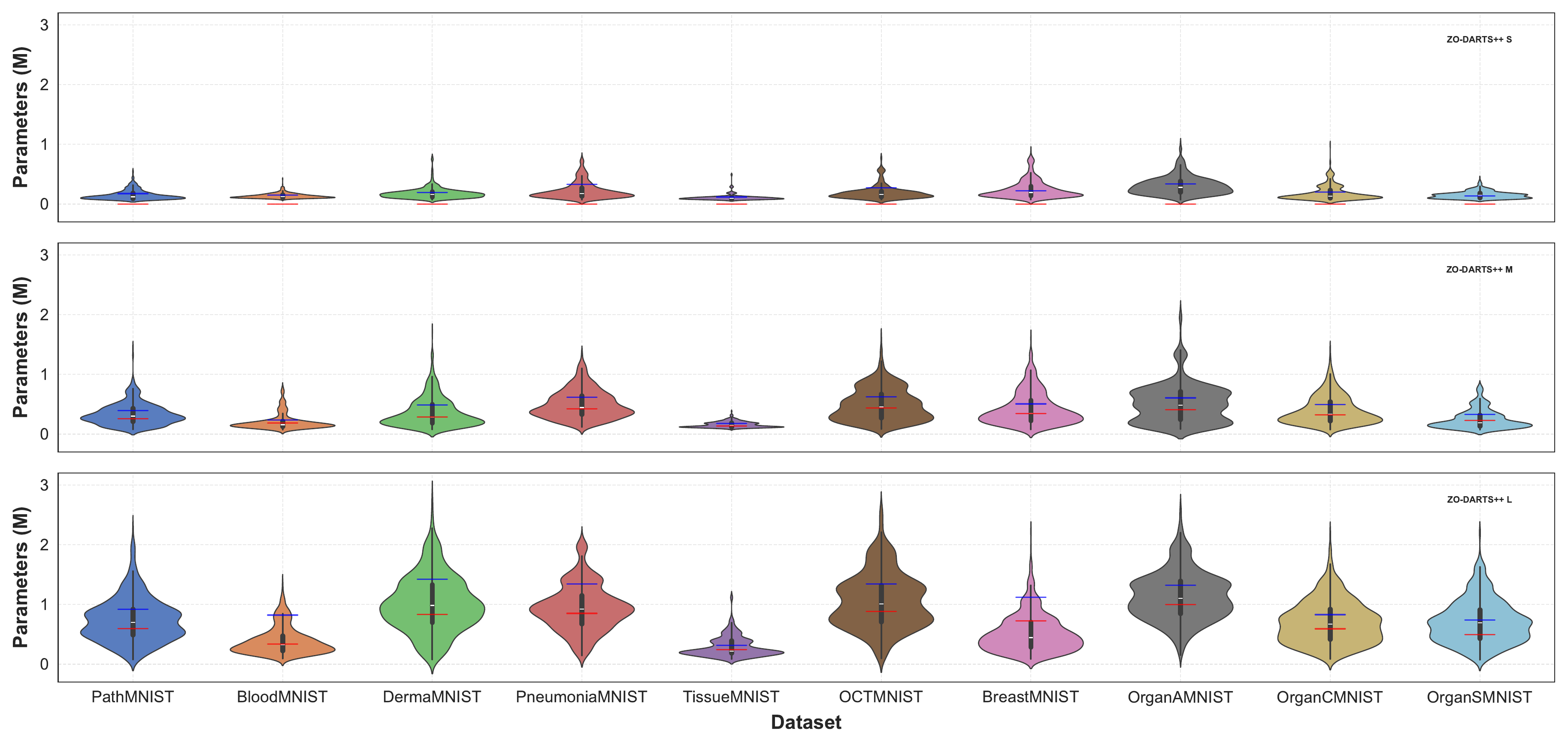}
    \caption{Violin plots of parameter numbers (million) distribution of models across various datasets. Each violin represents the range, density, and median parameter size for models sampled from supernets searched under three levels of constraints on the respective datasets. Blue and red lines represent the upper and lower bounds of constraints during the search.}
    \label{fig:size_constraint}
\end{figure*}

\subsection{Comparison under Resource Constraints}\label{sec:constraint}
Though the NAS algorithms are well developed, the adaptability to various platforms and resource constraints should be carefully considered. We restrict the upper and lower bounds of parameter consumption to limit the searched model size while ensuring basic performance. Facing these constraints, the algorithm will activate the size-variable search at the $20^{th}$ epoch. Table~\ref{tab:acc} also shows the performance of these three models on different datasets. The ZO-DARTS++ S model achieves accuracy comparable to that of ZO-DARTS++, while the M version shows a slight improvement. The largest model outperforms the trivial ZO-DARTS++ on eight datasets, with an average accuracy increase of nearly 1\%. Surprisingly, the smallest model outperforms the original unconstrained version on certain datasets. We attribute this to an unexpected regularization effect introduced by the tight constraints during the search process. Moreover, the accuracy of the constrained models increases with model size on five out of ten datasets, suggesting a potential correlation between model performance and the number of parameters. Furthermore, the average accuracy also improves as the model size increases, aligning with our intuition. Notably, there is only a slight time difference between the unconstrained ZO-DARTS++ and the constrained version, which we have omitted from the table for simplicity. In short, the conclusion that 38.6\% of the search time is saved on average compared to DARTS remains valid, showing the efficiency of our algorithm in constrained NAS problems. 

We also present the corresponding parameter counts of these sampled models in Table~\ref{tab:param}. Due to the unavailability of implementation details, we excluded the baseline methods AutoKeras and Google AutoML. Additionally, ResNet-18, with approximately 11.2 million parameters---two orders of magnitude larger than these NAS models---has a size that renders the comparison irrelevant, so we omitted it for simplicity. Notably, ZO-DARTS++ S achieves the smallest model size across nine out of ten datasets, outperforming all other methods. Its model scale is also much smaller than DARTS with the average parameter reduction at 35.9\%. To inspect the efficacy of ZO-DARTS++ clearer, we plotted the joint distribution of all the retrained models concerning both accuracy and number of parameters in Fig.\ref{fig:density}, which is a better interpretation of Table~\ref{tab:acc} and Table~\ref{tab:param}. The scatter plot records both model accuracy and parameter size. Each point represents a model sampled from the supernet during evaluation. To visually highlight the properties of different methods, we include two marginal distribution plots. Easily be seen, ZO-DARTS++ L takes a higher peak from the Y-axis, showing its better performance since it has various models with an accuracy greater than 90\%. Other models have very similar performance distributions. This aligns with the results presented in the previous tables. The X-axis plot highlights a significant variation in model sizes. Notably, a pronounced peak represents the concentration of smaller models from ZO-DARTS++ S, while two broader ridges correspond to the medium (M) and large (L) versions of ZO-DARTS++. In contrast, models from other methods exhibit a wider and less structured distribution of sizes. This observation supports the conclusion that the size-variable search scheme and the constraint-based search process effectively manage model size while maintaining strong performance.

Furthermore, as a verification of our algorithm, the distribution in Fig.~\ref{fig:size_constraint} is provided to depict the varying model sizes across different datasets. The violin plot of each level of the model on every dataset is generated via 900 models sampled from three searched supernets. Red and blue lines indicate the previous lower and upper-level size constraints used in the search. From ``S" to ``L", the model size distribution moves higher. And in most cases, our constraints are enclosed in the densest part. This observation shows the effectiveness of our size-variable search and the constraints used during the search.

In summary, these observations confirm that ZO-DARTS++ is both stable and feasible across different scenarios.

\subsection{Edge-Reliability in Differentiable NAS}\label{sec:reliability}

\begin{figure*}
    \centering
    \includegraphics[width=1\linewidth]{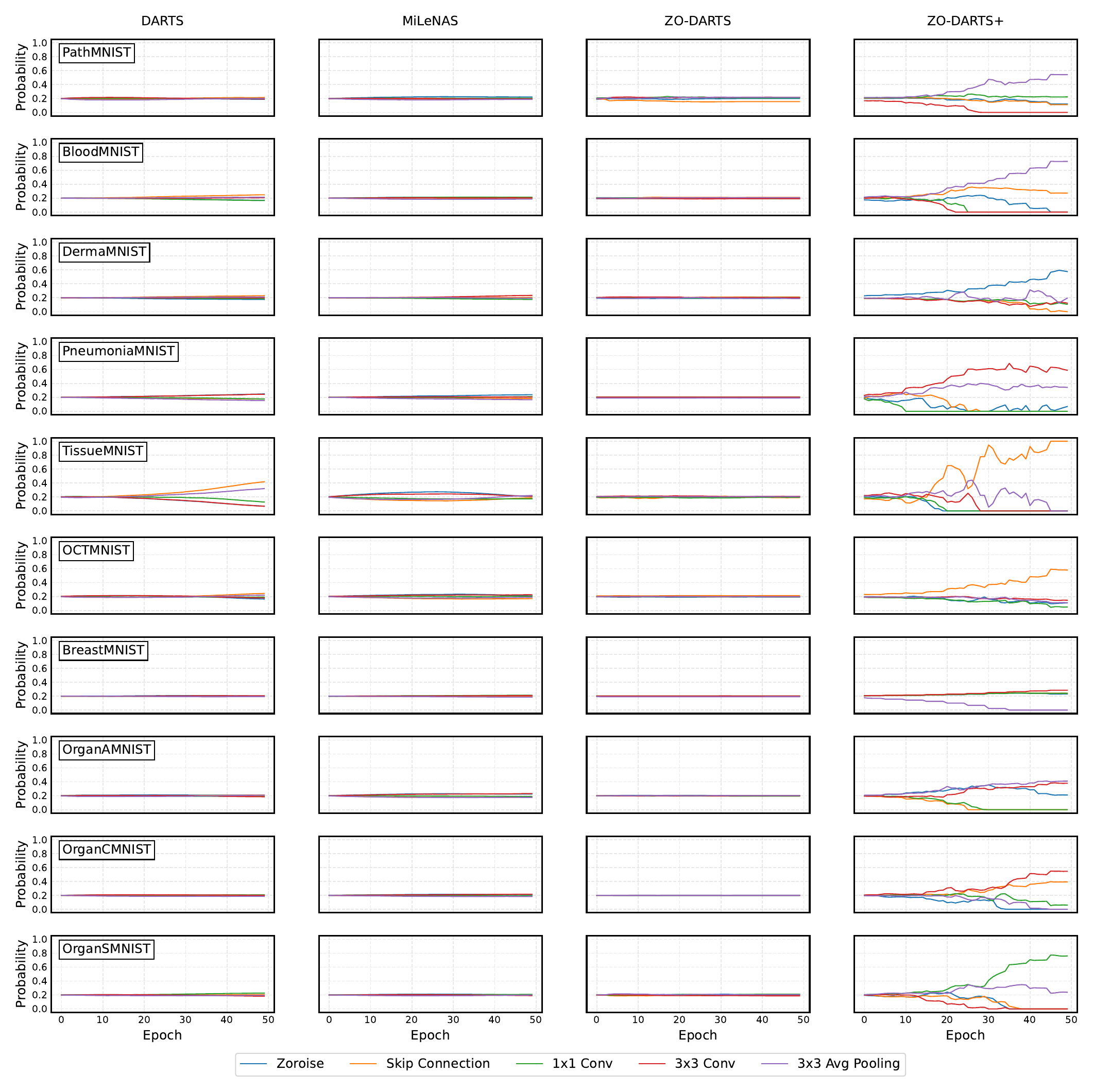}
    \caption{Probability rank variation of one edge during the search procedure. Lines with different colors represent different operations.}
    \label{fig:probabilities}
\end{figure*}
The probability values derived from architecture parameters are the most straightforward metric in determining the model structure for differentiable NAS methods. Commonly, the operation corresponding to the highest probability is chosen to build the final model. Unfortunately, long-term research has shown that there may exist many flaws. DARTS+~\cite{liang2019darts+} points out that probabilities of existing methods are close and may fall into non-learnable operations (e.g. skip connection) easily, which is also stressed by FairDARTS~\cite{chu2020fair}. With our approach, the sparsity-aware function can effectively mitigate this problem. We first analyze the progression of architecture probabilities across the two predecessors of ZO-DARTS++ and other differentiable NAS methods in Fig.\ref{fig:probabilities}. All models start optimization with the same initial architecture parameters, thus the same probability weights. 

\begin{figure*}[htbp]
    \centering
     \subfigure[Rankings of edges]{
        \includegraphics[width=0.48\textwidth]{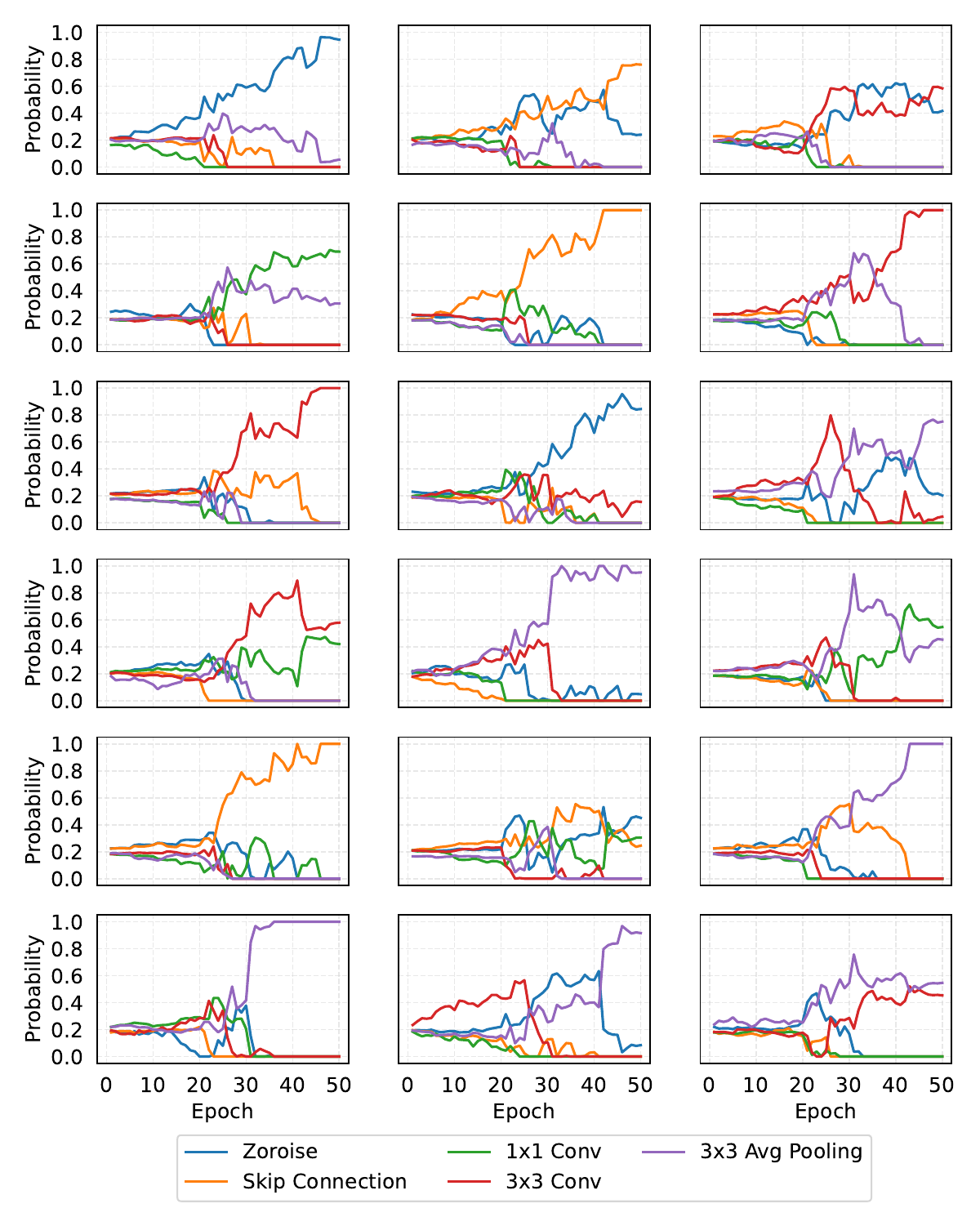}
        \label{fig:prob_edge}}
     \subfigure[Rankings of kernels]{
        \includegraphics[width=0.48\textwidth]{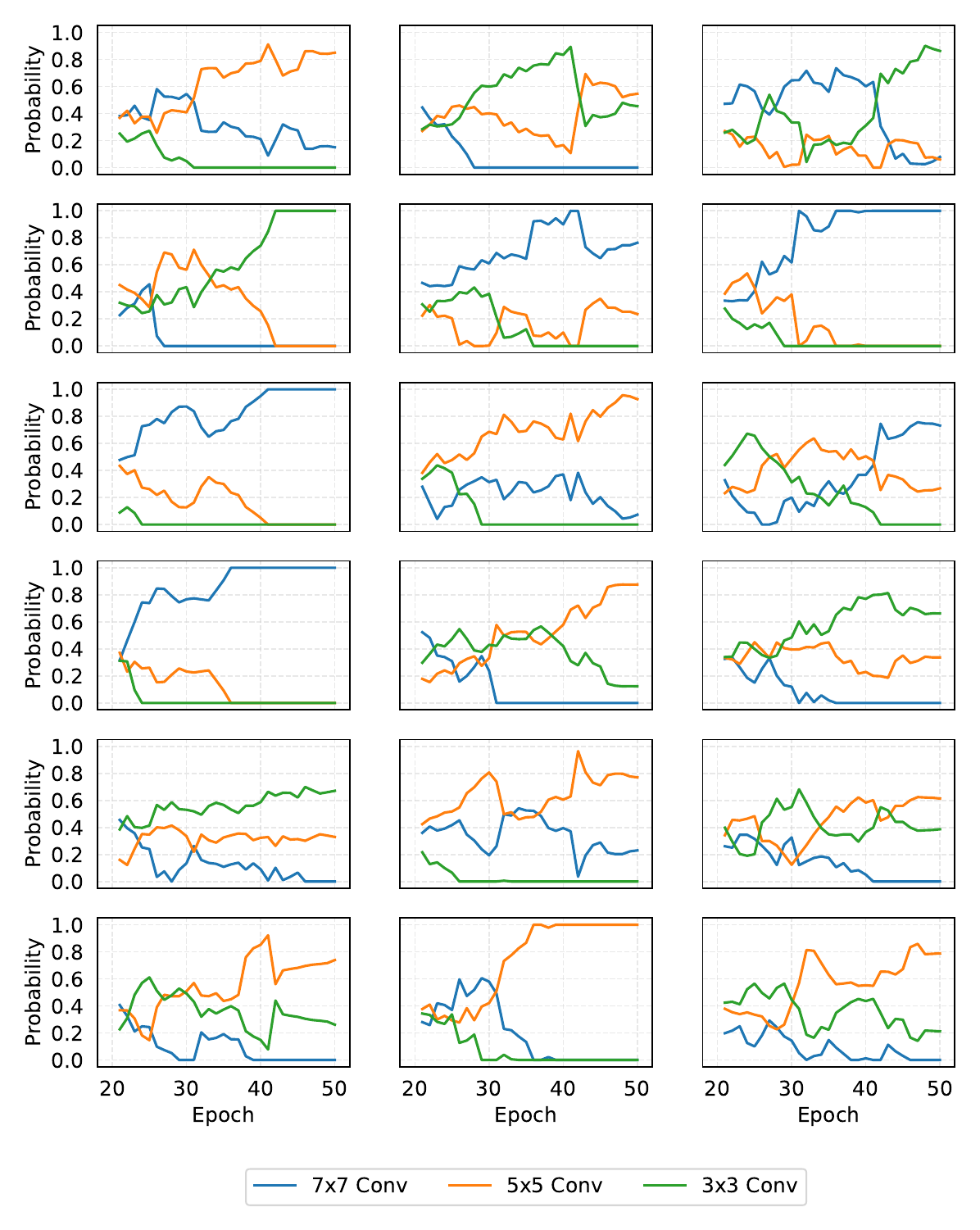}
        \label{fig:prob_kernel}}
    \caption{Probability rank variation of all edges during the search procedure on the OrganAMNIST dataset. Lines with different colors represent different operations or different sizes of kernels}\label{fig:prob_edge&kernel}
\end{figure*}

\begin{figure}
    \centering
    \includegraphics[width=0.48\textwidth]{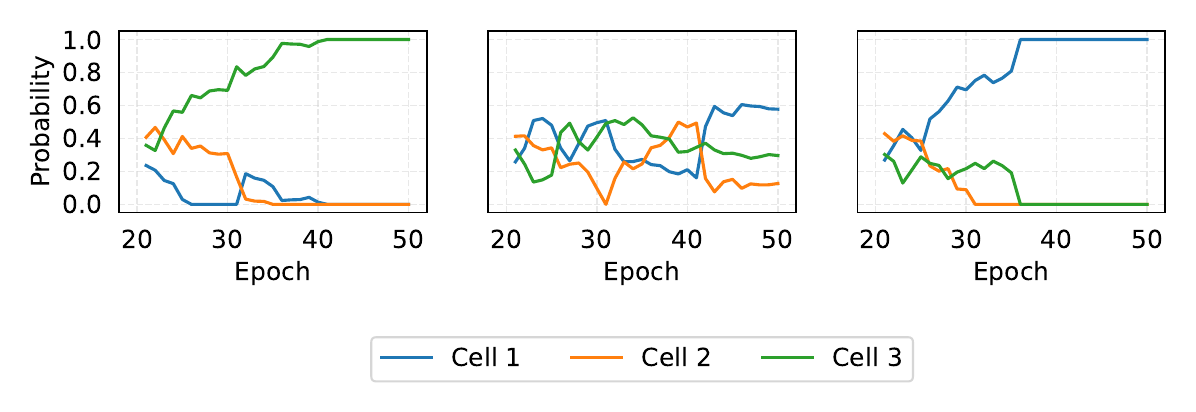}
    \caption{Probability rank variation of all depths during the search procedure. Lines with different colors represent different depths.}
    \label{fig:prob_depth}
\end{figure}

Throughout the search process, most methods exhibit limited variability in these weights, leading to challenges in reliably selecting the most effective operation---the probabilities remain too close together, allowing for easy rank changes, thus compromising the interpretability of operator selection. ZO-DARTS+ addresses this challenge by integrating sparsemax and an annealing strategy, enabling rapid convergence in selecting optimal operations. This innovation significantly improves performance while reducing the computational burden required to achieve clear probability separations. Building upon ZO-DARTS+, ZO-DARTS++ introduces several advancements, including a size-variable search scheme and an expanded architecture parameter space, enabling more flexible and effective searches. To evaluate the unique contributions of ZO-DARTS++, we plotted the probability distributions during the search process (using the OrganAMNIST dataset as a representative example). As shown in Fig.~\ref{fig:prob_edge}, ZO-DARTS++ demonstrates rapid convergence similar to ZO-DARTS+; however, its enhanced framework, driven by sparsity-aware functions and a more adaptable search space, ensures superior performance in both efficiency and model quality. This phenomenon is also verified on the probability plot of varied kernel sizes in Fig.~\ref{fig:prob_kernel} and varied depths in Fig.~\ref{fig:prob_depth}. Though we increase the search space of architecture probability parameters, they still present high divergence during optimization. That explains why we can stop the search after the $40^{th}$ epoch in our experiments. With the help of distinctive probability information, our sampling of structures rather than picking those with the highest probabilities is more significant. The architecture probabilities are sparse enough to represent a limited group of structures, and picking them out is basically the same as choosing the ``best'' model till the end of the search.

In addition, the analysis revealed distinct preferences for certain operations of one edge depending on the dataset. For example, in Fig.~\ref{fig:probabilities}, the 3x3 Conv operation was frequently selected for the PneumoniaMNIST dataset, whereas the 3x3 Avg Pooling was predominant in the OCTMNIST dataset, with other operations often excluded altogether from the final selection. This pattern highlights the adaptability of ZO-DARTS++ (and ZO-DARTS+), allowing it to effectively tailor its architecture to different types of data. These findings, coupled with robust accuracy results and reduced search times, confirm that ZO-DARTS++ is a highly validated model that demonstrates significant adaptability and reliability in its operation selection, making it a superior choice for medical image classification tasks.

\subsection{Comparison with POPNASv3}\label{sec:popnas}
We also compare the performance of ZO-DARTS++ with POPNASv3, a state-of-the-art NAS method that uses a Sequential Model-Based Optimization (SMBO) approach. Our experimental results, summarized in Tables~\ref{tab:acc},~\ref{tab:time}, and~\ref{tab:param}, demonstrate several distinct advantages of ZO-DARTS++ over POPNASv3.

Firstly, ZO-DARTS++ achieves significantly faster search times. As shown in Table~\ref{tab:time}, ZO-DARTS++ completes the search process in an average of 1,484.15 seconds across the MedMNIST datasets, while POPNASv3 requires an average of 31,283.87 seconds. This represents a reduction in computational time of approximately 95\%, highlighting the efficiency of our zeroth-order approximation technique in overcoming the computational challenges of the bi-level optimization while maintaining search accuracy.

Second, our sparsity-aware architecture generation, based on the sparsemax function, proves particularly effective in improving parameter efficiency. Traditional softmax-based approaches, such as those used in other differentiable NAS methods, tend to produce probabilities that do not sufficiently diverge between different operations. Although POPNASv3 uses a different approach (SMBO), it still faces the challenge of efficiently selecting operations that lead to smaller models without compromising performance. In contrast, our sparsemax-based method, enhanced with temperature annealing, pushes the architecture probabilities to more extreme values, resulting in a clearer selection of operations and the ability to construct smaller models.

This is evidenced by the ZO-DARTS++ S variant, which achieves remarkable efficiency with an average of only 0.137 million parameters while maintaining competitive accuracy (82.76\%), as shown in Table~\ref{tab:param}. Notably, this represents a better parameter efficiency than POPNASv3 average of 0.265 million parameters, as our models use fewer parameters while providing comparable or even superior performance on several datasets (Table~\ref{tab:acc}).

In addition, our size-variable search scheme, which dynamically adjusts kernel size and network depth during the search process, allows for more effective exploration of the architecture space compared to POPNASv3 fixed approach. This flexibility is particularly evident in the performance of our M and L variants. The ZO-DARTS++ L variant achieves the highest average accuracy of all tested approaches at 84.00\%, surpassing POPNASv3 average accuracy of 82.40\%. The L-variant shows particular strength on a variety of tasks, achieving 89.1\% accuracy on BreastMNIST, 94.2\% on OrganAMNIST , and 90.9\% on OrganCMNIST, as detailed in Table~\ref{tab:acc}.

The high performance across different model sizes demonstrates that our sparsemax-based operation selection process, combined with the kernel-variable convolution scheme and depth-variable architecture, creates a more robust and interpretable search framework. The ability to inherit weights from larger kernels and dynamically adjust the network depth allows ZO-DARTS++ to efficiently explore different architectural patterns while maintaining stable performance under varying resource constraints.

Overall, the comparison with POPNASv3 highlights the effectiveness of ZO-DARTS++ in delivering high-performance models with greater efficiency and adaptability to resource constraints. Our method not only speeds up the search process, but also achieves competitive or superior accuracy with fewer parameters, making it a compelling choice for real-world applications where computational resources are limited.

\section{Conclusion}\label{sec5}
In this article, we introduced the ZO-DARTS++ framework, which efficiently searches for high-performance architectures that adhere to specified resource constraints. The proposed approach addresses key challenges in the field of NAS, including low efficiency, opaque operation selection, and limited adaptability under varying resource conditions. By integrating the zeroth-order approximation for efficient gradient handling, the sparsemax function with temperature annealing for interpretable architecture distributions, and a size-variable search scheme for compact and accurate models, ZO-DARTS++ achieves a balance between model complexity and performance. Extensive evaluations on medical imaging datasets demonstrate the efficiency, flexibility, and superior performance of ZO-DARTS++ compared to state-of-the-art NAS methods. ZO-DARTS++ provides a robust and resource-aware framework for generating high-quality DL models tailored for real-world medical applications, paving the way for further research and practical deployment.

\section*{Acknowledgment}
This work is supported in part by the National Key R\&D Program of China under
grant 2022YFA1003900, in part by the NSFC under Grant 62231019, in part by the FAIR (Future Artificial Intelligence Research) project funded by the NextGenerationEU program within the PNRR-PE-AI scheme (M4C2, investment 1.3, line on Artificial Intelligence), and in part by the China Scholarship Council (CSC).

L. Xie extends his gratitude to D. Ardagna for the exceptional guidance as the host supervisor during his one-year doctoral visit in Milan, Italy. He also thanks E. Lomurno and M. Gambella for their critical input and collaboration on this work.

\ifCLASSOPTIONcaptionsoff
  \newpage
\fi



%\begin{thebibliography}{1}


\bibliographystyle{IEEEtran}
\bibliography{IEEEabrv,Bibliography}

\begin{thebibliography}{10}
\providecommand{\url}[1]{#1}
\csname url@rmstyle\endcsname
\providecommand{\newblock}{\relax}
\providecommand{\bibinfo}[2]{#2}
\providecommand\BIBentrySTDinterwordspacing{\spaceskip=0pt\relax}
\providecommand\BIBentryALTinterwordstretchfactor{4}
\providecommand\BIBentryALTinterwordspacing{\spaceskip=\fontdimen2\font plus
\BIBentryALTinterwordstretchfactor\fontdimen3\font minus \fontdimen4\font\relax}
\providecommand\BIBforeignlanguage[2]{{%
\expandafter\ifx\csname l@#1\endcsname\relax
\typeout{** WARNING: IEEEtran.bst: No hyphenation pattern has been}%
\typeout{** loaded for the language `#1'. Using the pattern for}%
\typeout{** the default language instead.}%
\else
\language=\csname l@#1\endcsname
\fi
#2}}

\bibitem{ramos2017detecting}
S.~Ramos, S.~Gehrig, P.~Pinggera, U.~Franke, and C.~Rother, ``Detecting unexpected obstacles for self-driving cars: Fusing deep learning and geometric modeling,'' in \emph{2017 IEEE Intelligent Vehicles Symposium (IV)}.\hskip 1em plus 0.5em minus 0.4em\relax IEEE, 2017, pp. 1025--1032.

\bibitem{zhang2021cross}
G.~Zhang, X.~Shen, Y.-D. Zhang, Y.~Luo, J.~Luo, D.~Zhu, H.~Yang, W.~Wang, B.~Zhao, and J.~Lu, ``Cross-modal prostate cancer segmentation via self-attention distillation,'' \emph{IEEE Journal of Biomedical and Health Informatics}, vol.~26, no.~11, pp. 5298--5309, 2021.

\bibitem{yan2018chitty}
R.~Yan, ``"chitty-chitty-chat bot": Deep learning for conversational ai.'' in \emph{IJCAI}, vol.~18, 2018, pp. 5520--5526.

\bibitem{2014Very}
K.~Simonyan and A.~Zisserman, ``Very deep convolutional networks for large-scale image recognition,'' \emph{Computer Science}, 2014.

\bibitem{he2016deep}
K.~He, X.~Zhang, S.~Ren, and J.~Sun, ``Deep residual learning for image recognition,'' in \emph{Proceedings of the IEEE conference on computer vision and pattern recognition}, 2016, pp. 770--778.

\bibitem{huang2017densely}
G.~Huang, Z.~Liu, L.~Van Der~Maaten, and K.~Q. Weinberger, ``Densely connected convolutional networks,'' in \emph{Proceedings of the IEEE conference on computer vision and pattern recognition}, 2017, pp. 4700--4708.

\bibitem{nasframework}
M.~Poyser and T.~P. Breckon, ``Neural architecture search: A contemporary literature review for computer vision applications,'' \emph{Pattern Recognition}, vol. 147, p. 110052, 2024.

\bibitem{liu2018darts}
H.~Liu, K.~Simonyan, and Y.~Yang, ``{DARTS:} differentiable architecture search,'' in \emph{ICLR}, 2019.

\bibitem{xiesnas}
S.~Xie, H.~Zheng, C.~Liu, and L.~Lin, ``Snas: stochastic neural architecture search,'' in \emph{International Conference on Learning Representations}, 2019.

\bibitem{dong2019searching}
X.~Dong and Y.~Yang, ``Searching for a robust neural architecture in four gpu hours,'' in \emph{Proceedings of the IEEE/CVF conference on computer vision and pattern recognition}, 2019, pp. 1761--1770.

\bibitem{dong2019one}
------, ``One-shot neural architecture search via self-evaluated template network,'' in \emph{Proceedings of the IEEE/CVF International Conference on Computer Vision}, 2019, pp. 3681--3690.

\bibitem{chu2020fair}
X.~Chu, T.~Zhou, B.~Zhang, and J.~Li, ``Fair darts: Eliminating unfair advantages in differentiable architecture search,'' in \emph{European conference on computer vision}.\hskip 1em plus 0.5em minus 0.4em\relax Springer, 2020, pp. 465--480.

\bibitem{zhang2021idarts}
M.~Zhang, S.~W. Su, S.~Pan, X.~Chang, E.~M. Abbasnejad, and R.~Haffari, ``idarts: Differentiable architecture search with stochastic implicit gradients,'' in \emph{ICML}, 2021, pp. 12\,557--12\,566.

\bibitem{tsaknakis2022implicit}
I.~Tsaknakis, P.~Khanduri, and M.~Hong, ``An implicit gradient-type method for linearly constrained bilevel problems,'' in \emph{ICASSP}, 2022, pp. 5438--5442.

\bibitem{liang2019darts+}
H.~Liang, S.~Zhang, J.~Sun, X.~He, W.~Huang, K.~Zhuang, and Z.~Li, ``Darts+: Improved differentiable architecture search with early stopping,'' \emph{arXiv preprint arXiv:1909.06035}, 2019.

\bibitem{xie2023zo}
L.~Xie, K.~Huang, F.~Xu, and Q.~Shi, ``{ZO-DARTS}: Differentiable architecture search with zeroth-order approximation,'' in \emph{ICASSP}.\hskip 1em plus 0.5em minus 0.4em\relax IEEE, 2023.

\bibitem{jin2019rc}
X.~Jin, J.~Wang, J.~Slocum, M.-H. Yang, S.~Dai, S.~Yan, and J.~Feng, ``Rc-darts: Resource constrained differentiable architecture search,'' \emph{arXiv preprint arXiv:1912.12814}, 2019.

\bibitem{zhang2020data}
X.~Zhang, J.~Chang, Y.~Guo, G.~Meng, S.~Xiang, Z.~Lin, and C.~Pan, ``Data: Differentiable architecture approximation with distribution guided sampling,'' \emph{IEEE Transactions on Pattern Analysis and Machine Intelligence}, vol.~43, no.~9, pp. 2905--2920, 2020.

\bibitem{chen2021progressive}
X.~Chen, L.~Xie, J.~Wu, and Q.~Tian, ``Progressive darts: Bridging the optimization gap for nas in the wild,'' \emph{International Journal of Computer Vision}, vol. 129, pp. 638--655, 2021.

\bibitem{xu2019pc}
Y.~Xu, L.~Xie, X.~Zhang, X.~Chen, G.-J. Qi, Q.~Tian, and H.~Xiong, ``Pc-darts: Partial channel connections for memory-efficient architecture search,'' in \emph{ICLR}, 2019.

\bibitem{lorraine2020optimizing}
J.~Lorraine, P.~Vicol, and D.~Duvenaud, ``Optimizing millions of hyperparameters by implicit differentiation,'' in \emph{AISTATS}, 2020, pp. 1540--1552.

\bibitem{xieefficient}
L.~Xie, E.~Lomurno, M.~Gambella, D.~Ardagna, M.~Roveri, M.~Matteucci, and Q.~Shi, ``An efficient neural architecture search model for medical image classification,'' in \emph{ESANN}, 2024.

\bibitem{falanti2023popnasv3}
A.~Falanti, E.~Lomurno, D.~Ardagna, and M.~Matteucci, ``Popnasv3: a pareto-optimal neural architecture search solution for image and time series classification,'' \emph{Applied Soft Computing}, vol. 145, p. 110555, 2023.

\bibitem{gambella_cnas_2022}
\BIBentryALTinterwordspacing
M.~Gambella, A.~Falcetta, and M.~Roveri, ``\BIBforeignlanguage{en}{{CNAS}: {Constrained} {Neural} {Architecture} {Search}},'' in \emph{\BIBforeignlanguage{en}{{SMC}}}.\hskip 1em plus 0.5em minus 0.4em\relax IEEE, 2022. [Online]. Available: \url{https://ieeexplore.ieee.org/document/9945080/}
\BIBentrySTDinterwordspacing

\bibitem{li2022lc}
G.~Li, M.~Xu, S.~Giancola, A.~Thabet, and B.~Ghanem, ``Lc-nas: Latency constrained neural architecture search for point cloud networks,'' in \emph{2022 International Conference on 3D Vision (3DV)}.\hskip 1em plus 0.5em minus 0.4em\relax IEEE, 2022, pp. 1--11.

\bibitem{lomurno2024pomonag}
E.~Lomurno, S.~Mariani, M.~Monti, and M.~Matteucci, ``Pomonag: Pareto-optimal many-objective neural architecture generator,'' \emph{arXiv preprint arXiv:2409.20447}, 2024.

\bibitem{2024ReCNAS}
C.~Peng, Y.~Li, R.~Shang, and L.~Jiao, ``Recnas: Resource-constrained neural architecture search based on differentiable annealing and dynamic pruning,'' \emph{IEEE transactions on neural networks and learning systems}, no.~2, p.~35, 2024.

\bibitem{caionce}
H.~Cai, C.~Gan, T.~Wang, Z.~Zhang, and S.~Han, ``Once-for-all: Train one network and specialize it for efficient deployment,'' in \emph{International Conference on Learning Representations}, 2020.

\bibitem{lu2021neural}
Z.~Lu, G.~Sreekumar, E.~Goodman, W.~Banzhaf, K.~Deb, and V.~N. Boddeti, ``Neural architecture transfer,'' \emph{IEEE transactions on pattern analysis and machine intelligence}, vol.~43, no.~9, pp. 2971--2989, 2021.

\bibitem{sarti2023neural}
S.~Sarti, E.~Lomurno, and M.~Matteucci, ``Neural architecture transfer 2: A paradigm for improving efficiency in multi-objective neural architecture search,'' \emph{arXiv preprint arXiv:2307.00960}, 2023.

\bibitem{he2020milenas}
C.~He, H.~Ye, L.~Shen, and T.~Zhang, ``Milenas: Efficient neural architecture search via mixed-level reformulation,'' in \emph{CVPR}, 2020.

\bibitem{nesterov2017random}
Y.~Nesterov and V.~Spokoiny, ``Random gradient-free minimization of convex functions,'' \emph{FoCM}, vol.~17, no.~2, pp. 527--566, 2017.

\bibitem{caiproxylessnas}
H.~Cai, L.~Zhu, and S.~Han, ``Proxylessnas: Direct neural architecture search on target task and hardware,'' in \emph{International Conference on Learning Representations}, 2019.

\bibitem{martins2016softmax}
A.~Martins and R.~Astudillo, ``From softmax to sparsemax: A sparse model of attention and multi-label classification,'' in \emph{ICML}.\hskip 1em plus 0.5em minus 0.4em\relax PMLR, 2016.

\bibitem{medmnistv2}
J.~Yang, R.~Shi, D.~Wei, Z.~Liu, L.~Zhao, B.~Ke, H.~Pfister, and B.~Ni, ``Medmnist v2-a large-scale lightweight benchmark for 2d and 3d biomedical image classification,'' \emph{Scientific Data}, 2023.

\bibitem{JMLR:v24:20-1355}
\BIBentryALTinterwordspacing
H.~Jin, F.~Chollet, Q.~Song, and X.~Hu, ``Autokeras: An automl library for deep learning,'' \emph{Journal of Machine Learning Research}, vol.~24, no.~6, pp. 1--6, 2023. [Online]. Available: \url{http://jmlr.org/papers/v24/20-1355.html}
\BIBentrySTDinterwordspacing

\bibitem{google_automl}
{Google AutoML Vision}, \url{https://cloud.google.com/automl}.

\bibitem{dong2020bench}
X.~Dong and Y.~Yang, ``Nas-bench-201: Extending the scope of reproducible neural architecture search,'' in \emph{ICLR}, 2020.

\end{thebibliography}

\vfill


\end{document}